\documentclass[twoside]{article}

\usepackage[accepted]{cameraready/aistats2024}

\usepackage[round]{natbib}

\usepackage{minitoc}
\usepackage{microtype}
\usepackage{graphicx}
\usepackage{subcaption}
\usepackage{booktabs}
\usepackage{pifont}
\usepackage{tabularx}
\usepackage[dvipsnames,svgnames]{xcolor}
\usepackage{pgfplots}
\newcommand{\smallblacksquare}{\scalebox{0.5}{$\blacksquare$}}

\usepackage{hyperref}

\usepackage[toc,page,header]{appendix}
\usepackage{minitoc}

\usepackage{amsmath}
\usepackage{amssymb}
\usepackage{mathtools}
\usepackage{amsthm}
\usepackage{bbm}
\usepackage{enumitem}
\definecolor{bblue}{HTML}{4F81BD}
\definecolor{rred}{HTML}{C0504D}
\definecolor{ggreen}{HTML}{9BBB59}
\definecolor{ppurple}{HTML}{9F4C7C}
\usepackage[capitalize,noabbrev]{cleveref}
\usepackage{adjustbox}
\usepackage{tikz}
\usepackage{algorithm}
\usepackage{algorithmic}
\usetikzlibrary{calc,math}
\usetikzlibrary{decorations}
\usetikzlibrary{decorations.markings}
\usetikzlibrary{shapes.geometric}
\usetikzlibrary{arrows}
\pgfdeclarelayer{bg}
\pgfsetlayers{bg,main}

\usepackage{pgfplots}
\usepackage[framemethod=TikZ]{mdframed}
\usepackage{comment}
\usepackage{fontawesome}
\usepackage{pgfplotstable}
\usepackage{colortbl}
\usepackage{wrapfig}
\mdfsetup{%
    middlelinecolor =   none,
    middlelinewidth =   1pt,
    backgroundcolor =   blue!3,
    roundcorner     =   5pt,
}

\theoremstyle{plain}
\newtheorem{theorem}{Theorem}[section]
\newtheorem{proposition}[theorem]{Proposition}

\theoremstyle{definition}
\newtheorem{definition}[theorem]{Definition}

\theoremstyle{remark}

\usepackage[textsize=tiny]{todonotes}

\newcommand{\Dtrain}{\mathcal{D}_{\mathrm{train}}}
\newcommand{\Dtest}{\mathcal{D}_{\mathrm{test}}}

\newcommand{\point}[1]{\ding{117} {\em #1}\enskip}

\newcolumntype{C}{>{\centering\arraybackslash}X}

\everypar{\looseness=-1}
\hypersetup{%
    colorlinks = true,
    citecolor  = blue,
    linkcolor  = blue,
}

\begin{document}
\runningtitle{DAGnosis: Localized Identification of Data Inconsistencies using Structures}
\runningauthor{Huynh*, Berrevoets*, Seedat, Crabbé, Qian, van der Schaar}

\twocolumn[

\aistatstitle{DAGnosis: Localized Identification of Data Inconsistencies using Structures}

\aistatsauthor{ Nicolas Huynh* \And Jeroen Berrevoets* \And  Nabeel Seedat}
\aistatsaddress{ University of Cambridge \And University of Cambridge \And University of Cambridge} 
\aistatsauthor{Jonathan Crabbé \And Zhaozhi Qian \And Mihaela van der Schaar}
\aistatsaddress{  University of Cambridge \And University of Cambridge \And University of Cambridge}]

\begin{abstract}
   Identification and appropriate handling of inconsistencies in data at deployment time is crucial to reliably use machine learning models. While recent data-centric methods are able to identify such inconsistencies with respect to the training set, they suffer from two key limitations: (1) suboptimality in settings where features exhibit statistical independencies, due to their usage of compressive representations and (2) lack of localization to pin-point why a sample might be flagged as inconsistent, which is important to guide future data collection. We solve these two fundamental limitations using directed acyclic graphs (DAGs) to encode the training set's features probability distribution and independencies as a structure. Our method, called DAGnosis, leverages these structural interactions to bring valuable and insightful data-centric conclusions. DAGnosis unlocks the localization of the causes of inconsistencies on a DAG, an aspect overlooked by previous approaches. Moreover, we show empirically that leveraging these interactions (1) leads to more accurate conclusions in detecting inconsistencies, as well as (2) provides more detailed insights into why some samples are flagged.
\end{abstract}

\vfill

\everypar{\looseness=-1}
\linepenalty=1000
\
\section{INTRODUCTION} \label{sec:introduction}

{\bf No Data, No Machine Learning.}  Data plays a crucial role in machine learning as it is used to train and test models \citep{park2021reliable,jain2020overview, Sambasivan}. To ensure reliable downstream performance, it is essential to have structured mechanisms to assess new data in relation to our training data \citep{Seedat2022DataSUITEDI, saria2019tutorial}. This is a critical concern which should be addressed as neglecting it may lead to poor downstream performance for models evaluated on incongruous samples  \citep{10.1145/3035918.3054782,renggli2021data}. 
This consideration motivates recent interest in {\it data-centric AI} (DCAI) \citep{liang2022advances,seedat2022dc},
which aims to develop ``systematic methods to evaluate [\ldots], the data used to train and test the AI model'' \citep{liang2022advances}. Building such data-centric methods confers the immediate advantage of flexibility, as insights about the data can be applied to any downstream model.

{\bf Inconsistencies.} A key challenge in DCAI is to flag inconsistencies in new data with respect to the training set. Inconsistencies can manifest in real-world settings for a variety of reasons. Even if the new samples are in-distribution, they may exhibit inconsistencies due to finite-sample effects exacerbated by regions of low data coverage (e.g. underrepresented sub-groups) \citep{Krawczyk2016LearningFI, yuksekgonul2023beyond}, or data biases \citep{Torralba2011UnbiasedLA}  in the training dataset. 
The identification of these inconsistencies is of paramount importance to ensure reliable downstream performance and can guide future data collection. It justifies a systematic and principled data-centric approach, leading to {\it rich and valuable insights}.

{\bf Tabular Data and Sparse Connections.} Of particular interest in this paper is tabular data, which is ubiquitous in real-world and high-stake settings, such as medicine, finance or economics \citep{Borisov2021DeepNN}. 

Data-SUITE \citep{Seedat2022DataSUITEDI} is the method most relevant to our work, as it flags inconsistencies in the tabular domain. It computes feature-wise uncertainty, in the form of prediction intervals. One key element of the approach is the input to obtain these intervals, which is a compressive representation (e.g. PCA or autoencoder) of the {\it complete} input. However, using such compressive representations is suboptimal for two reasons. First, it overlooks the sparse dependencies between features in tabular datasets  \citep{yang22i,kalisch2007estimating} (which differ from images or text where features are very tightly coupled). In such datasets, not all the features might be relevant \citep{jordon2018knockoffgan}. Second, it does not permit the \textit{localization} of the reasons why a sample is deemed inconsistent, which is problematic from an auditing perspective.
We discuss this in more details in \cref{sec:related}.
\begin{figure}[t]
    \begin{tikzpicture}[
        var/.style = {rounded corners, fill=black!10},
        vert/.style={circle, thick, draw=black, inner sep=0, minimum size=2mm},scale = 1
    ]
        \node (new_sample_start) at (-2.3,0) {};
        \node (new_sample_end) at (2.3, 0) {};
        \foreach \x in {1, 2, 3, 4, 5, 6} {
            \node[var] at ($(new_sample_start)!{\x/7}!(new_sample_end) $) {$x_\x$};
        }
        \node at ($(new_sample_start)!.5!(new_sample_end) + (0, .5)$) {new sample};
        \node (center_graph) at ($(new_sample_start) + (-1, -1.4)$) {};
        \path let \p1 = (center_graph) in coordinate (others) at (-\x1,\y1);
        \node (center_datasuite) at ($(new_sample_end) + (-1, -.3)$) {};

        \node (others_start) at ($(others) + (-.5, 1.1)$) {};
        \node (others_end) at ($(others) + (-.5, -.8)$) {};
        \begin{scope}
            \foreach \x in {1, 2, 3, 4, 5, 6} {
                \node[var, fill=DarkGreen!25, inner sep=1.5, rounded corners=.5mm] at ($(others_start)!{\x/7}!(others_end)$) {\tiny $x_\x$};
                \node[anchor=west, align=left] at ($(others_start)!{\x/7}!(others_end) + (.3, .01)$) {\textcolor{DarkGreen}{\tiny normal}};
                
            }
        \end{scope}

        \begin{scope}
            \node[vert, fill=red!25, label={[label distance=-1mm]\tiny $x_1$}] (x1) at ($(center_graph) + .5*(0, 1)$) {};
            \node[vert, fill=DarkGreen!25, label={[label distance=-1mm]\tiny $x_2$}] (x2) at ($(center_graph) + .5*(0, 0)$) {};
            \node[vert, fill=DarkGreen!25, label={[label distance=-1mm]\tiny $x_3$}] (x3) at ($(center_graph) + .5*(0.951, 0.309)$) {};
            \node[vert, fill=DarkGreen!25, label={[label distance=-1mm]\tiny $x_4$}] (x4) at ($(center_graph) + .5*(0.587, -0.809)$) {};
            \node[vert, fill=DarkGreen!25, label={[xshift=-1mm, label distance=-1mm] \tiny $x_5$}] (x5) at ($(center_graph) + .5*(-0.587, -0.809)$) {};
            \node[vert, fill=DarkGreen!25, label={[label distance=-1mm]north:\tiny $x_6$}] (x6) at ($(center_graph) + .5*(-0.95, 0.309)$) {};
    
            \draw[-latex, thick] (x5) -- (x2);
            \draw[-latex, thick] (x6) -- (x2);
            \draw[-latex, thick] (x2) -- (x3);
            \draw[-latex, thick] (x3) -- (x1);
        \end{scope}

        \node[anchor= west, text width=2cm, align=flush left] (dagnosis_text) at ($(center_graph) + (.7, 0)$) {\tiny\baselineskip=4pt\textcolor{DarkBlue}{$x_1$ is \textcolor{Maroon}{abnormal} given $x_3$, and $x_2$ is \textcolor{DarkGreen}{normal} given $x_6$, $x_5$ and $x_3$}\par};
        \node[align=right] at ($(dagnosis_text) + (0, .7)$) {\textcolor{DarkBlue}{DAGnosis}};
        
        \path let \p1 = (dagnosis_text) in coordinate (others_text) at (-\x1,\y1);
        \node[align=left] at ($(others_text) + (0, .7)$) {Others};
        \node[anchor= east, text width=2cm, align=flush left]  at ($(others) - (.7, 0)$) {\tiny\baselineskip=4pt{$x_1$ is not detected as abnormal because the sample is {\it mostly} \textcolor{DarkGreen}{normal}}\par};

        \draw[->, DarkGreen, very thick] ($(new_sample_start) + (0, .2)$) node[circle, fill=white, draw=DarkGreen, inner sep=0, minimum size=1mm, very thick] {} to[out=135, in=90, looseness=1.5] node[rotate=-45, rounded corners=.5mm, fill=white, draw=none, midway] {\small evaluate} ($(center_graph) + (0, 1.1)$);

        \draw[->, DarkGreen, very thick] ($(new_sample_end) + (0, .2)$) node[circle, fill=white, draw=DarkGreen, inner sep=0, minimum size=1mm, very thick] {} to[out=45, in=90, looseness=1.5] node[rotate=45, rounded corners=.5mm, fill=white, draw=none, midway] {\small evaluate} ($(others) + (0, 1.1)$);

        \begin{pgfonlayer}{bg}
            \draw[rounded corners, fill=blue!5, draw=none] ($(center_graph) + (-1, 1)$) rectangle ($(dagnosis_text) + (1.2, -.7)$);

            \draw[rounded corners, fill=black!5, draw=none] ($(others) + (1, 1)$) rectangle ($(others_text) + (-1.2, -.7)$);
        \end{pgfonlayer}
        
    \end{tikzpicture}

    \caption{\textbf{DAGnosis Provides {\em Precise} Analysis.} DAGnosis takes a radically different approach compared to other data-evaluation methods. Rather than evaluating each dimension of a new sample in relation to all the other dimensions, we evaluate in relation to {\it the structure} of the sample. This may lead to different samples being flagged while giving interpretation for that conclusion.}
    \label{fig:pull}
\end{figure}
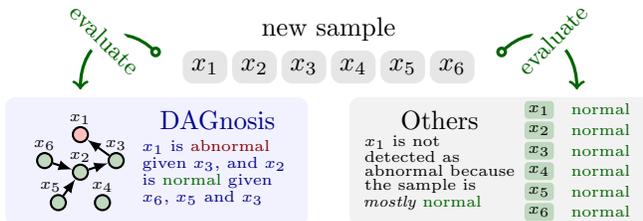

 To address these limitations, we present DAGnosis, a data-centric evaluation strategy for the tabular domain. {\it DAGnosis addresses the problem of flagging inconsistencies,} for which it requires two components: a way to leverage sparsity and independence in the tabular data, where we leverage structures modeled as {\it directed acyclic graphs} (DAGs); and a way to flag instances in the data, where we build on {\it conformal prediction}. DAGnosis provides {\it localized} instance-wise conclusions: having flagged an inconsistency, it gives a set of features which explain it. Localization is important because it can inform future data collection (enriching the {\it training set}) or suggest new measurements of features (correcting the {\it test samples} which exhibit measurement noise).
 In \Cref{fig:pull} we see that DAGnosis crucially relies on a Bayesian network describing the features, which makes it \emph{interpretable by design}~\citep{BARREDOARRIETA202082}. Unlike previous data-centric methods, our conclusions take into account the interactions between features through the structure encoded in a DAG.  

{\bf Contributions.} DAGnosis advances the state-of-the-art as follows: \textbf{\textcolor{ForestGreen}{\textcircled{1}}}  \textbf{Conceptually}, DAGnosis identifies and addresses the suboptimality of compressive representations. To the best of our knowledge, DAGnosis is the first method to leverage structures for data-centric insights. It unlocks the localization of the reasons why a sample is deemed inconsistent, an aspect overlooked by the state-of-the-art. \textbf{\textcolor{ForestGreen}{\textcircled{2}}} 
 \textbf{Technically}, DAGnosis learns a DAG describing the relationships between the features and trains feature-wise conformal predictors. It conditions them on relevant variables as determined by the DAG. \textbf{\textcolor{ForestGreen}{\textcircled{3}}}  \textbf{Empirically}, we demonstrate in \cref{sec:experiments} that DAGnosis outperforms the SOTA on accuracy of inconsistency detection and downstream accuracy when deferring predictions on inconsistent samples. Furthermore, we provide a detailed case study on a real-world dataset in \cref{walkthrough}, showing how practitioners can benefit from DAGnosis to gain understanding of inconsistencies.

\vspace{3mm}

\section{RELATED WORK} \label{sec:related}

\vspace{3mm}

{\bf Data-centric Evaluation.} Even though data-centric insights are important, they have been mostly neglected in favor of model-dependent conclusions. This is epitomized by the field of predictive uncertainty quantification \citep{Gawlikowski2021ASO}, where the idea is to categorize samples with respect to the uncertainty in the prediction of a given model (e.g. with Gaussian Processes \citep{Rasmussen2003GaussianPF}, Bayesian Neural Networks \citep{ghosh2018structured} or using ensembles \citep{Lakshminarayanan2016SimpleAS}). 

In this work, we move away from this branch and instead give conclusions with respect to the data itself.

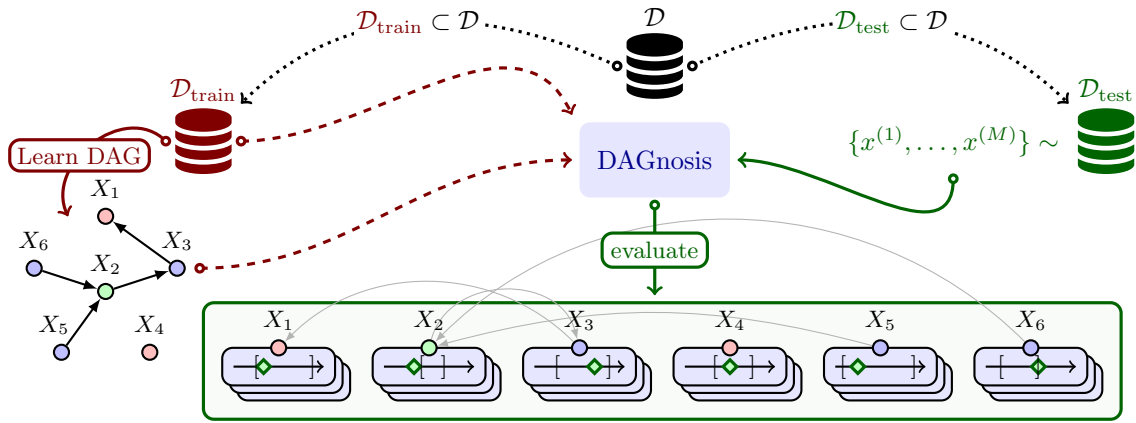
\begin{figure*}[ht!]

    \begin{tikzpicture}[
        vert/.style={circle, thick, draw=black, inner sep=0, minimum size=2mm},
        sample/.style={diamond, fill=green!25, inner sep=.5mm, draw=DarkGreen, very thick},
        scale=1, transform shape
    ]

        \node[label={[label distance=-1.5mm]above:\textcolor{Maroon}{$\mathcal{D}_\text{train}$}}] (train) at (0,0) {\textcolor{Maroon}{\Huge \faDatabase}};
        \node[label={[label distance=-1.5mm]above:\textcolor{DarkGreen}{$\mathcal{D}_\text{test}$}}] (test) at ($(train) + (12,0)$) {\textcolor{DarkGreen}{\Huge \faDatabase}};
        \node[label={[label distance=-1.5mm]above:$\mathcal{D}$}] (data) at ($(train)!.5!(test) + (0, 1)$) {\Huge \faDatabase};
        
        \draw[->, dotted, very thick] ($(data) + (-.5, 0)$) node[circle, very thick, solid, fill=white, inner sep=0, minimum size=1mm, draw=black] {} to[out=155, in=45] node[midway, fill=white, inner sep=1] {$\textcolor{Maroon}{\mathcal{D}_\text{train}}\subset\mathcal{D}$} (train);
        \draw[->, dotted, very thick] ($(data) + (.5, 0)$) node[circle, very thick, solid, fill=white, inner sep=0, minimum size=1mm, draw=black] {} to[out=25, in=135] node[midway, fill=white, inner sep=1] {$\textcolor{DarkGreen}{\mathcal{D}_\text{test}}\subset\mathcal{D}$} (test);

        \node[anchor=east] (query) at ($(test) + (-.5, 0)$) {\textcolor{DarkGreen}{$\{x^{(1)}, \dots,x^{(M)}\} \sim$}};

        \node[inner sep=4mm] (center_graph) at ($(train) + (-1.3, -2)$) {};
        \node (center_dagnosis) at ($(train)!.5!(test) + (0, -0.25)$) {};
    
        \draw[rounded corners, fill=blue!10, draw=none] ($(center_dagnosis) + (-1, .5)$) rectangle ($(center_dagnosis) + (1,-.5)$);
        \node at (center_dagnosis) {\textcolor{DarkBlue}{DAGnosis}};

        \begin{scope}
            \node[vert, fill=red!25, label={\small $X_1$}] (x1) at ($(center_graph) + (0, 1)$) {};
            \node[vert, fill=green!25, label={\small $X_2$}] (x2) at ($(center_graph) + (0, 0)$) {};
            \node[vert, fill=blue!25, label={\small $X_3$}] (x3) at ($(center_graph) + (0.951, 0.309)$) {};
            \node[vert, fill=red!25, label={\small $X_4$}] (x4) at ($(center_graph) + (0.587, -0.809)$) {};
            \node[vert, fill=blue!25, label={[xshift=-1mm] \small $X_5$}] (x5) at ($(center_graph) + (-0.587, -0.809)$) {};
            \node[vert, fill=blue!25, label={north:\small $X_6$}] (x6) at ($(center_graph) + (-0.95, 0.309)$) {};
    
            \draw[-latex, thick] (x5) -- (x2);
            \draw[-latex, thick] (x6) -- (x2);
            \draw[-latex, thick] (x2) -- (x3);
            \draw[-latex, thick] (x3) -- (x1);
        \end{scope}
    
        \draw[->, very thick, Maroon] ($(train) + (-.5, 0)$) node[circle, draw=Maroon, fill=white, very thick, inner sep=0, minimum size=1mm] {} to[out=135, in=115, looseness=1.2] node[pos=.6, xshift=-.2, inner sep=1mm, fill=white, draw=Maroon, very thick, rounded corners] {\textcolor{Maroon}{\small Learn DAG}} ($(x1)!.5!(x2) + (-.5, .5)$);
    
        \draw[->, very thick, Maroon, dashed] ($(train) + (.5, 0)$) node[circle, solid, draw=Maroon, fill=white, very thick, inner sep=0, minimum size=1mm] {} to[out=0, in=115] ($(center_dagnosis) + (-1.1,.6)$);
    
        \draw[->, very thick, Maroon, dashed] ($(x3) + (.3, 0)$) node[circle, solid, draw=Maroon, fill=white, very thick, inner sep=0, minimum size=1mm] {} to[out=0, in=180] ($(center_dagnosis)+ (-1.1, 0)$);
    
        \draw[->, DarkGreen, very thick] ($(query) + (0, -.5)$) node[circle, solid, draw=DarkGreen, fill=white, very thick, inner sep=0, minimum size=1mm] {} to[out=270, in=0] ($(center_dagnosis) + (1.1, 0)$);

        \node (output) at ($(center_dagnosis) + (0,-1.5)$) {};
        \node[vert, fill=red!25, label={\small $X_1$}] (out_1) at ($(output) + (-5, -1)$) {};
        \node[vert, fill=green!25, label={\small $X_2$}] (out_2) at ($(output) + (-3, -1)$) {};
        \node[vert, fill=blue!25, label={\small $X_3$}] (out_3) at ($(output) + (-1, -1)$) {};
        \node[vert, fill=red!25, label={\small $X_4$}] (out_4) at ($(output) + (1, -1)$) {};
        \node[vert, fill=blue!25, label={\small $X_5$}] (out_5) at ($(output) + (3, -1)$) {};
        \node[vert, fill=blue!25, label={\small $X_6$}] (out_6) at ($(output) + (5, -1)$) {};
    
        \foreach \x [count=\xi] in {out_1, out_2, out_3, out_4, out_5, out_6} {
            \draw[->, thick] ($(\x) + (-.6, -.25)$) -- ($(\x) + (.6, -.25)$);
        }
    
        \node[sample] at ($(out_1) + (-.2, -.25)$) {};
        \node at ($(out_1) + (-.3, -.25) $) {\small $[$};
        \node at ($(out_1) + (.45, -.25)$) {\small $]$};
    
        \node[sample] at ($(out_2) + (-.2, -.25)$) {};
        \node at ($(out_2) + (-.1, -.25) $) {\small $[$};
        \node at ($(out_2) + (.2, -.25)$) {\small $]$};
    
        \node[sample] at ($(out_3) + (.2, -.25)$) {};
        \node at ($(out_3) + (-.3, -.25) $) {\small $[$};
        \node at ($(out_3) + (.4, -.25)$) {\small $]$};
    
        \node[sample] at ($(out_4) + (-.0, -.25)$) {};
        \node at ($(out_4) + (-.2, -.25) $) {\small $[$};
        \node at ($(out_4) + (.3, -.25)$) {\small $]$};
    
        \node[sample] at ($(out_5) + (-.3, -.25)$) {};
        \node at ($(out_5) + (-.5, -.25) $) {\small $[$};
        \node at ($(out_5) + (.45, -.25)$) {\small $]$};
    
        \node[sample] at ($(out_6) + (.1, -.25)$) {};
        \node at ($(out_6) + (-.4, -.25) $) {\small $[$};
        \node at ($(out_6) + (.1, -.25)$) {\small $]$};

        \begin{pgfonlayer}{bg}
    
            \draw[rounded corners, fill=DarkGreen!2.5, draw=DarkGreen, very thick] ($(out_1) + (-1, .6)$) rectangle ($(out_6) + (.95, -.7) + (.25, -.25)$);
            
            \draw[-latex, black!30] (out_3) to[out=135, in=45] (out_1);
            \draw[-latex, black!30] (out_6) to[out=135, in=45] (out_2);
            \draw[-latex, black!30] (out_5) to[out=165, in=15] (out_2);
            \draw[-latex, black!30] (out_2) to[out=75, in=105, looseness=1.2] (out_3);

            \foreach \x in {out_1, out_2, out_3, out_4, out_5, out_6} {
                \draw[fill=blue!10, draw=black, thick, rounded corners] ($(\x) + (-.75, 0) + (.2, -.2)$) rectangle ($(\x) + (.75, -.5) + (.2, -.2)$);
            }
            
            \foreach \x in {out_1, out_2, out_3, out_4, out_5, out_6} {
                \draw[fill=blue!10, draw=black, thick, rounded corners] ($(\x) + (-.75, 0) + (.1, -.1)$) rectangle ($(\x) + (.75, -.5) + (.1, -.1)$);
            }
    
            \foreach \x in {out_1, out_2, out_3, out_4, out_5, out_6} {
                \draw[fill=blue!10, draw=black, thick, rounded corners] ($(\x) + (-.75, 0)$) rectangle ($(\x) + (.75, -.5)$);
            }
    
        \end{pgfonlayer}

        \draw[->, very thick, DarkGreen] ($(center_dagnosis) + (0, -.6)$) node[circle, solid, draw=DarkGreen, fill=white, very thick, inner sep=0, minimum size=1mm] {} -- node[very thick, draw=DarkGreen, fill=white, rounded corners, pos=.5] {\small \textcolor{DarkGreen}{evaluate}} ($(output) + (0, -.3)$);

    \end{tikzpicture}

    \caption{{\bf High-level Overview of DAGnosis.} DAGnosis evaluates samples in a test-bed dataset \textcolor{DarkGreen}{$\mathcal{D}_\text{test}$}. It first learns a DAG (using a variety of structure learners). Next, DAGnosis builds prediction intervals for every feature using conformal prediction. They are conditioned on smart subsets of the data's features, informed by the DAG.} \label{fig:dagnosis}
\end{figure*}
As such, Data-SUITE \citep{Seedat2022DataSUITEDI} is the method most relevant to our work. However, it has several key limitations:
\vspace{3mm}
\begin{mdframed}

\textbf{\em (i) non-adaptiveness of the conditioning sets.} The same input (i.e. the representation obtained with PCA) is given to $d$ feature regressors, while in practice each feature might depend on a different set of variables. DAGnosis creates a set of conditioning variables specific to each feature. In this way, it is more flexible with respect to the specificity of each feature. \vspace{3mm}

\textbf{\em (ii) localization.} Data-SUITE does not offer a localized explanation on why examples are flagged as inconsistent in terms of the input features themselves; because the features are combined in the compressive representation, it is difficult to contextualise an inconsistent feature as resulting from an abnormal value conditioned on the other features. DAGnosis unlocks localization, and provides important insights which can guide future data collection. \vspace{3mm}

\textbf{\em (iii) sparse interactions between the features.} Sparse interactions are ubiquitous in real settings, evidenced by abundant noise variables \citep{kalisch2007estimating}. In such settings, using a compressive representation leads to a loss of information, affecting the quality of the conformal predictors, and hence the data-centric conclusions.  

\end{mdframed}

\vspace{-3mm}

{\bf Structure.} In order to account for the sparse interactions between features in the tabular domain --- an aspect overlooked by Data-SUITE --- DAGnosis leverages structures (DAGs).  A DAG consists of vertices and edges, where the vertices represent the random variables comprising a feature set, and the edges model direct dependence \citep{Koller2009ProbabilisticGM, guyon2007causal, berrevoets2023causal}. We term the setting {\it sparse} \citep{ng2020role} when the number of edges is low. It is these sparse settings that we hope to model more accurately with DAGnosis.

DAGs can be discovered, via a variety of structure learners \citep{Zheng2018DAGsWN, Peters2017ElementsOC,verma1990,verma1990causal,geiger1994learning,berrevoets23dstruct}, or instead be provided or completed by the user, when prior knowledge is available \citep{hasan2022kcrl,sinha2021using, Sachs2005CausalPN}.

\section{DAGNOSIS: IDENTIFYING INCONSISTENCIES USING STRUCTURES} \label{method}
We are interested in scenarios where we have a training dataset and want to flag inconsistent test samples without relying on a downstream model. Moreover, we wish to go beyond the current data-centric capabilities of  {\it just} flagging samples. We also want to provide a {\it reason or localization} as to why they were flagged. This is important to make data-centric methods principled from an auditing perspective.

{\bf Data.} We consider a $d$-dimensional feature space $\mathcal{X} \subseteq \mathbb{R}^{d}$, where we have access to a training dataset, $\mathcal{D}_{\mathrm{train}} = \{x^{k} \mid k \in [n_{\mathrm{train}}]\} \subset \mathcal{X}$, and a test set $\mathcal{D}_{\mathrm{test}} = \{x^{j} \mid j \in [n_{\mathrm{test}}]\} \subset \mathcal{X}$, with $n_{\mathrm{train}}$ and $n_{\mathrm{test}}$ being the cardinalities of $\mathcal{D}_{\mathrm{train}}$ and $\mathcal{D}_{\mathrm{test}}$, respectively. Moreover, we assume $\mathcal{D}_{\mathrm{train}}$ is composed of i.i.d. samples coming from a distribution $P^{*}$. We denote indices of features with subscripts, i.e. $x_i$ is the $i$-th feature of $x$. When $\mathcal{S}$ is a set of indices, $x_{\mathcal{S}}$ denotes the restriction of $x$ to the indices in $\mathcal{S}$. For future convenience, we also define $\mathcal{I}(\mathcal{E})$ which returns the feature-indices present in the set of random variables $\mathcal{\mathcal{E}}$.

As shown in \cref{fig:dagnosis}, we wish to flag inconsistencies in $\mathcal{D}_{\mathrm{test}}$ with respect to $\mathcal{D}_{\mathrm{train}}$. In order to characterize inconsistencies in $\mathcal{D}_\mathrm{test}$ at deployment time, we aim to provide feature-wise conclusions for every sample $x = [x_1, \dots, x_d]^\top \in \mathcal{D}_\mathrm{test}$, which will be aggregated into sample-wise conclusions. These conclusions inform whether or not a sample is labeled inconsistent.

{\bf Structures Representing Data.} A key contribution of our work is to approach the problem by leveraging structures. The structures of interest are directed acyclic graphs (DAG). Intuitively these structures act as a compact representation of a factorization of $P^{*}$ \cite[Chapter 3 \& 4]{Koller2009ProbabilisticGM}. 

Let $\mathcal{G} = (\mathcal{V}, \mathcal{E})$ denote a DAG, comprised of a set of vertices ($\mathcal{V}$) and edges ($\mathcal{E}$), with $\mathcal{E} \subset \mathcal{V} \times \mathcal{V}$. If $(V_{i}, V_{j}) \in  \mathcal{E}$ then $(V_{j}, V_{i}) \notin  \mathcal{E}$, making $(V_i, V_j) \neq (V_j, V_i)$. In practice, we consider Bayesian networks (BN) describing $\mathbf{X}$, a random vector following the training distribution $P^{*}$, i.e. $\mathbf{X} = (X_{V})_{V \in \mathcal{V}} \in \mathcal{X}$  is a random vector with its coordinates in correspondence with the graph's vertices. The whole point of using Bayesian networks is that they encode  (conditional) independencies between the features graphically, with the notion of d-separation \citep{Geiger2013dSeparationFT}. As a direct consequence, for each feature $X_i$, we can identify a minimal set of features to best regress $X_i$, according to the BN. We first recall the definition of Markov blankets.

\begin{definition}[Markov blanket] \label{def:markov_blanket}
Let $\mathcal{S} = \{X_1, X_2, ..., X_d\}$ be a set of random variables. For $i \in [d]$, a Markov blanket of the random variable $X_{i}$ in $\mathcal{S}$ is any subset $\mathcal{S}_{i} \subseteq	\mathcal{S}$ such that: $$X_{i} \perp\!\!\!\!\perp (\mathcal{S} \setminus \mathcal{S}_{i}) \mid \mathcal{S}_i$$ A minimal Markov blanket is called a Markov boundary.
\end{definition}

\begin{definition}[Markov boundary] \label{def:markov_boundary}
A Markov boundary of a random variable $X_i$ in a set $\mathcal{S} \coloneqq \{X_1, \dots, X_d\}$ is any subset $\mathcal{S}^- \subset \mathcal{S}$ which is a Markov blanket (\cref{def:markov_blanket}), but does not contain any proper subset which itself is a Markov blanket of $X_i$. We will denote the Markov boundary of $X_i$ as $\mathcal{X}^-(X_i)$.
\end{definition}

A Markov boundary enables us to find the minimal set of variables which capture all the {\it sufficient} information to describe a particular feature. Moreover, we can find them graphically in a Bayesian network \citep{Koller2009ProbabilisticGM}. As we will describe in \cref{sec:dagnosis}, we use these minimal sets to build conditional conformal predictors.  Crucially, this contrasts using an {\it entire} representation, likely containing irrelevant information, especially in tabular settings with many irrelevant variables (i.e. a sparse setting).

\subsection{Structure-based Assessment of Samples}
\label{sec:dagnosis}

\begin{algorithm*}[t]
\caption{ {\bf -- Training.} We describe how we train each feature-specific conformal predictor, informed by the DAG. While we consider $\mathcal{G}$ as an input to the training routing, $\mathcal{G}$ can also be learned prior to this. }
\label{alg:train_cp}
\begin{algorithmic}
\STATE \textbf{Input:} A proper training set $\mathcal{D}_{\mathrm{train}}^{+}$, a calibration set $\mathcal{D}_{\mathrm{cal}}$, significance levels $\alpha$, $\alpha_{lo}$ , $\alpha_{hi}$, and a DAG $\mathcal{G}$
\STATE \textbf{Output:} A list of conformal predictors $\{(l_{i, \alpha}, r_{i, \alpha}) : i \in [d]\}$
\FOR{$i \in [d]$}
\STATE $\mathcal{D}_i \leftarrow \{ (f_i(x), x_i) : \forall x \in \mathcal{D}_{\mathrm{train}}^{+}, \forall i \in [d] \}$ \hfill\COMMENT{Construct the training set using $\mathcal{G}$}
\STATE $\hat{q}_{i, \alpha_{lo}}$\texttt{.train($\alpha_{lo}$,$\mathcal{D}_{i}$)} and $\hat{q}_{i, \alpha_{hi}}$\texttt{.train($\alpha_{hi}$,$\mathcal{D}_{i}$)} \hfill\COMMENT{Fit the quantile regressors}
\STATE $\epsilon_{i, \alpha} \leftarrow$\texttt{calibrate($\hat{q}_{i, \alpha_{lo}}, \hat{q}_{i, \alpha_{hi}}, \mathcal{D}_\mathrm{cal}$)} \hfill\COMMENT{Calibrate the quantile regressors}
\STATE $l_{i, \alpha} \leftarrow \hat{q}_{i, \alpha_{lo}} - \epsilon_{\alpha, i}$ and $r_{i, \alpha} \leftarrow \hat{q}_{i, \alpha_{hi}} + \epsilon_{\alpha, i}$ \hfill\COMMENT{Define the conformal predictors} 
\STATE Add $l_{i, \alpha}$ and $r_{i, \alpha}$ to the list of conformal predictors
\ENDFOR
\end{algorithmic}
\end{algorithm*}

To flag inconsistencies in the data, DAGnosis models feature-wise uncertainty in a frequentist setting using a graphical representation of $\mathcal{D}_{\mathrm{train}}$. Our desideratum is to obtain distribution-free prediction intervals (PIs) for each feature, with a coverage guarantee (in order to control a desired False Positive Rate when flagging inconsistencies). To fulfill this desideratum, DAGnosis leverages conformal prediction \citep{vovk2005conformal}. Specifically, for all $i \in [d]$ and any $x \in \mathcal{D}_{\mathrm{test}}$ as well a significance level $\alpha \in (0,1)$,  we construct PIs $[l_{i, \alpha}(x), r_{i, \alpha}(x)]$ for 
 $x_i$. Given $\alpha$, conformal prediction comes with a marginal coverage guarantee stemming from a calibration step, when the exchangeability assumption is satisfied \citep{Balasubramanian2014ConformalPF}.  More details are provided in Appendix~\ref{appendix:cp}.

Rather than directly using the complete $x$ as input to the conformal estimators, we will exploit the structure given by a discovered DAG $\mathcal{G}$--- which models conditional dependencies ---to come up with more informed PIs in a compact and accurate way. 
As we will confirm in \cref{sec:experiments}, doing so leads to more accurate discovery, as well as improved downstream task performance when deferring prediction on samples flagged as inconsistent.

\textbf{Structures and Independence.} Consider the autoregressive factorization $P(\mathbf{X}) = P(X_1)P(X_2 | X_1)...P(X_d|X_1,...,X_{d-1})$ over $d$ random variables, which holds true for any distribution $P$.
With this factorization and no further assumptions on (conditional) independencies, one can identify the simple Markov boundary, $\mathcal{X}^-(X_i)=\{X_1,..., X_d\} \setminus \{X_i\}$, which amounts to using all the other variables to describe $X_{i}$. However, in many settings, variables exhibit (conditional) independence relationships, yet the approach we have just described (essentially taken by other benchmarks) does not account for it. This insight motivates the use of structures.

We design our method to be agnostic to the way the structure is provided. It can come from a structure learner, which takes as input a dataset $\mathcal{D}$, and outputs a DAG $\mathcal{G}$; or can be given a priori. As such, we can choose any structure learner in the wide range of conditional independence testing based (CIT), like the PC algorithm \citep{spirtes2000causation}, or score-based methods. In some settings \citep{Sachs2005CausalPN,mooij2016,pinna2013reconstruction}, we can leverage prior knowledge and provide (or complete) the underlying ground-truth DAG. In our experiments, we assume no access to prior knowledge and thus learn the DAG ourselves, in light of fair comparison to other benchmark methods.

{\bf Constructing Feature-wise Prediction Intervals.} Given a DAG $\mathcal{G}$ as input, we compute adaptive prediction intervals for each feature $X_i$, and any sample $x \in \mathcal{X}$, denoted by $[l_{i, \alpha}(x), r_{i, \alpha}(x)]$, where $\alpha$ is the significance level. We use Conformalized Quantile Regression (CQR) \cite{Romano2019ConformalizedQR} as our inductive conformal prediction method, which has been shown to outperform other inductive conformal prediction benchmarks. To perform conformal prediction, we split $\mathcal{D}_{\mathrm{train}}$ into a training set $\mathcal{D}_{\mathrm{train}}^{+}$ and a calibration set $\mathcal{D}_{\mathrm{cal}}$. We now describe the construction of the feature-wise prediction intervals. The following steps are conducted for each feature $i \in [d]$:

\point{\bf Step 1} Given two significance levels $\alpha_{lo}, \alpha_{hi}$, train conditional quantile regressors $\hat{q}_{i, \alpha_{lo}}, \hat{q}_{i, \alpha_{hi}}$, using  $\mathcal{D}_{\mathrm{train}}^{+}$. The input to the quantile regressors for feature $i$ is denoted by $f_i(x)$. The quantile regressors are trained using the $\alpha_{lo}$ and $\alpha_{hi}$-pinball losses.

Crucial to DAGnosis is our (graph-based) definition of $f_i$, which is $f_i(x) \coloneqq x_{\mathcal{I}(\mathcal{X}^-(X_i))}$.
This definition directly leverages the structure $\mathcal{G}$ to define $f_i$. Alternatively stated, $f_{i}(x)$ contains all the feature-indices present in $X_i$'s Markov boundary. This is in stark contrast to methods which use a compressing representation (PCA/Autoencoder) where $x_i$ could indirectly be present.

\point{\bf Step 2} Use a calibration set  $\mathcal{D}_{\mathrm{cal}}$ to  compute a set of non-conformity scores $C_i = \{ E_{i,j} \mid j \in [\lvert \mathcal{D}_{\mathrm{cal}} \rvert] \}$, and compute the $(1-\alpha)(1+ \frac{1}{\lvert \mathcal{D}_{\mathrm{cal}} \rvert})$ empirical quantile of $C_i$, denoted as $\epsilon_{\alpha, i}$ (see  \cref{appendix:cp} for details). The prediction intervals for a particular sample $x$ and a feature $i \in [d]$ are then given by $[\hat{q}_{i, \alpha_{lo}}(f_{i}(x)) - \epsilon_{\alpha, i},  \hat{q}_{i, \alpha_{hi}}(f_{i}(x)) + \epsilon_{\alpha, i}]$. We summarize the training part of DAGnosis in Algorithm~\ref{alg:train_cp}.

\textit{Remark:} DAGnosis does not require learning a full probabilistic model of the data. It solely needs a structure before performing feature-wise CP. CP is a distribution-free paradigm, hence DAGnosis makes no specific distributional assumption.

\subsection{Leveraging Structure to Flag Inconsistencies}
In the previous section we detailed how DAGnosis builds feature-wise conformal predictors. We can now use these predictors to flag inconsistencies in new test examples with respect to the training set.

{\bf Benefits of Using a Structure.}
Using a structure in DAGnosis directly solves problems related to other data-centric methods when attempting to flag inconsistent examples. Indeed, (1) for every feature, the structure defines  which set of conditioning variables we should use; (2) by virtue of having an explicit set of conditioning variables, we localize the causes of the inconsistencies; (3) we account for irrelevant variables, which are isolated nodes in the structure, and which are not used as input to the conformal prediction step. 

{\bf Metrics.} Similar to \cite{Seedat2022DataSUITEDI}, the conformal prediction framework brings an interesting conclusion at the sample-level, which is whether or not each feature value falls inside its associated confidence interval. This is encapsulated in the notion of inconsistency.
\begin{definition}[Inconsistency] 
Let $x \in \mathcal{D}_{\mathrm{test}}$ be a test instance for which we construct a $(1-\alpha)$-PI, $[l_{i, \alpha}(x), r_{i, \alpha}(x)]$, for each feature $x_i , i \in [d]$, with a significance level $\alpha \in (0,1)$. For each $x_i , i \in [d]$, the \emph{feature inconsistency} is a binary variable indicating if $x_i$ falls out of the PI. 
\begin{align}
     \nu_i(x) \equiv  \mathbb{I} (x_{i} \notin [l_{i, \alpha}(x), r_{i, \alpha}(x)])
\end{align}
The \emph{instance inconsistency} $\nu(x)$ is obtained by summing the feature inconsistencies $\nu_i(x)$, that is
    $\nu(x) = \sum_{i=1}^{d} \nu_i (x) $.
A sample $x$ is inconsistent if $\nu(x) > 0$.
\label{def:incons}
\end{definition}
With this definition, we can now give conclusions on the data \textit{itself}, summarized in \cref{alg:test_cp} in Appendix \ref{appendix:inference}.

\textit{Remark}: 
We compare DAGnosis to conventional out-of-distribution detectors in Appendix \ref{app:ood}, highlighting that DAGnosis can flag in-distribution inconsistencies like Data-SUITE \citep{Seedat2022DataSUITEDI}.

\section{EXPERIMENTS} \label{sec:experiments}

In this section, we show that taking into account structures leads to \textit{accurate} and \textit{localized} detection of inconsistencies. We also demonstrate how DAGnosis enables reliable downstream performance. For details on the experimental setup \footnote{\url{https://github.com/nicolashuynh/DAGNOSIS}} \footnote{\url{https://github.com/vanderschaarlab/DAGNOSIS}}, please refer to \cref{app:exp_details}. 

{\bf Baselines.} For the rest of this section, we consider DAGnosis and the baseline Data-SUITE as it is the only comparable method. We consider different variants of DAGnosis, where the nature of the DAG used by DAGnosis differs among the variants: \textit{GT} (ground-truth DAG describing $P^{*}$, when it is known), \textit{Autoregressive} (autoregressive factorization, described in \cref{sec:dagnosis}), \textit{NOTEARS} (NT) (\cite{Zheng2020LearningSN}) and \textit{DAGMA} (\cite{bello2022dagma}), two differentiable structure learners, and \textit{PC} ( \cite{spirtes2000causation}), a constraint-based structure learner. For Data-SUITE, we use $\frac{d}{2}$ as the representation dimension similarly to \cite{Seedat2022DataSUITEDI}, and use CQR as the Inductive Conformal Prediction (ICP) method, since it is considered the SOTA for ICP.

\subsection{DAGnosis Flags Inconsistencies Accurately} \label{inconsistency}

\begin{table*}[ht]
	\centering
	\normalsize
        \setlength{\tabcolsep}{2pt}
	\caption{{\bf Results on Inconsistency Detection.} We report the $F_1$ score ($\uparrow$), precision (prec.) ($\uparrow$) and recall (rec.) ($\uparrow$) for the inconsistency detection task over different settings with decreasing sparsity (higher $s$ indicates less sparse).  Mean and $1.96 \times$ standard errors are reported.  We benchmark against DAGnosis paired with a naive DAG modeling the autoregressive factorization (\textbf{\em DN Auto}), Data-SUITE (\textbf{\em DS CQR}), and have \colorbox{green!10}{colored} our method's (\textbf{\em DN NT}) row for clarity. Beyond the relevant benchmarks, we have also included DAGnosis with a ground truth DAG \textbf{\em (DN GT)}. This acts as an oracle ``upper bound'' and is \colorbox{black!5}{shaded} to clearly distinguish it from the other methods. Our results show that DAGnosis (using NOTEARS) improves $F_1$ scores, precision, and recall.}
	
        \vspace{0.2em}
	
        \label{tab:exp1}
	\begin{tabularx}{\textwidth}{r | *{3}{C} | *{3}{C} | *{3}{C} | *{3}{C}}
			\toprule
			 & \multicolumn{3}{c|}{$s= 10$}
			 & \multicolumn{3}{c|}{$s= 20$}
			 & \multicolumn{3}{c|}{$s= 30$}
			 & \multicolumn{3}{c}{$s= 40$} \\
			 
			 & 
                {\scriptsize $F_1$} {\tiny ($\uparrow$)} & {\scriptsize Prec.} {\tiny ($\uparrow$)} & {\scriptsize Rec.} {\tiny ($\uparrow$)} & 
                {\scriptsize $F_1$} {\tiny ($\uparrow$)} & {\scriptsize Prec.} {\tiny ($\uparrow$)} & {\scriptsize Rec.} {\tiny ($\uparrow$)} & 
                {\scriptsize $F_1$} {\tiny ($\uparrow$)} & {\scriptsize Prec.} {\tiny ($\uparrow$)} & {\scriptsize Rec.} {\tiny ($\uparrow$)} & 
                {\scriptsize $F_1$} {\tiny ($\uparrow$)} & {\scriptsize Prec.} {\tiny ($\uparrow$)} & {\scriptsize Rec.} {\tiny ($\uparrow$)} \\
			 
			\toprule
			
			& \multicolumn{12}{c}{\it Linear SEMs} \\
			\toprule

			\small DN Auto&
			{\scriptsize		 0.81} {\tiny (.03)}& 
			{\scriptsize	   0.89} {\tiny (.01)}&  
			{\scriptsize		 0.76} {\tiny (.04)}&
			{\scriptsize		 0.84} {\tiny (.03)}& 
			{\scriptsize		 0.89} {\tiny (.01)}& 
			{\scriptsize		 0.79} {\tiny (.04)}&
			{\scriptsize		 0.86} {\tiny (.03)}& 
			{\scriptsize		 \textbf{0.91}} {\tiny (.01)}& 
			{\scriptsize		 0.82} {\tiny (.04)}&
			  {\scriptsize		 0.86} {\tiny (.02)}& 
			  {\scriptsize		 0.91} {\tiny (.01)}&  
			  {\scriptsize		 0.83} {\tiny (.04)}\\
			  
			\small DS CQR &  
			{\scriptsize	 0.81} {\tiny (.03)}& 
			{\scriptsize	 0.88} {\tiny (.01)}& 
			{\scriptsize	 0.75} {\tiny (.04)}& 
			
			{\scriptsize	 0.84} {\tiny (.03)}& 
			{\scriptsize	 0.90} {\tiny (.01)}& 
			{\scriptsize	 0.80} {\tiny (.04)}&

			{\scriptsize	 0.85} {\tiny (.03)}& 
			{\scriptsize	 0.90} {\tiny (.01)}& 
			{\scriptsize	 0.81} {\tiny (.04)}&
			 
			{\scriptsize	 0.85} {\tiny (.02)}& 
			{\scriptsize	 0.91} {\tiny (.01)}& 
			{\scriptsize	 0.80} {\tiny (.04)}\\
			
			\rowcolor{green!10}  \small DN NT & 
			{\scriptsize	 \textbf{0.85}} {\tiny (.02)}& 
			{\scriptsize	 \textbf{0.89}} {\tiny (.01)}& 
			{\scriptsize	 \textbf{0.82}} {\tiny (.04)}&
			  
			{\scriptsize	 \textbf{0.88}} {\tiny (.02)}& 
			{\scriptsize	 \textbf{0.90}} {\tiny (.01)}& 
			{\scriptsize	 \textbf{0.86}} {\tiny (.03)}&

			{\scriptsize	 \textbf{0.87}} {\tiny (.03)}& 
			{\scriptsize	 0.90} {\tiny (.01)}&  
			{\scriptsize	 \textbf{0.85}} {\tiny (.04)}&

			{\scriptsize	 \textbf{0.88}} {\tiny (.02)}&
			{\scriptsize	 \textbf{0.91}} {\tiny (.01)}&
			{\scriptsize	 \textbf{0.85}} {\tiny (.04)} \\
			
			\midrule
			\rowcolor{black!5} \small (DN GT) & 
			{\scriptsize	 {0.85}} {\tiny (.02)}& 
			{\scriptsize	 {0.90}} {\tiny (.01)}& 
			{\scriptsize	 {0.82}} {\tiny (.04)}&
			  
			{\scriptsize	 {0.88}} {\tiny (.02)}& 
			{\scriptsize	 {0.90}} {\tiny (.01)}& 
			{\scriptsize	 {0.86}} {\tiny (.03)}&

			{\scriptsize	 {0.87}} {\tiny (.02)}& 
			{\scriptsize	 {0.90}} {\tiny (.01)}&  
			{\scriptsize	 {0.85}} {\tiny (.04)}&

			{\scriptsize	 0.88} {\tiny (.02)}&
			{\scriptsize	 0.91} {\tiny (.01)}&
			{\scriptsize	 0.85} {\tiny (.04)} \\

			\toprule
			
			& \multicolumn{12}{c}{\it MLP SEMs} \\
			 
			\toprule

			\small DN Auto &
			{\scriptsize	 0.78} {\tiny (.07)} & 
			{\scriptsize	 0.89} {\tiny (.02)} &  
			{\scriptsize	 0.72} {\tiny (.09)} &
			  
			{\scriptsize	 0.83} {\tiny (.08)} & 
			{\scriptsize	 0.89} {\tiny (.03)} & 
			{\scriptsize	 0.81} {\tiny (.11)} &

			{\scriptsize	 0.79} {\tiny (.06)} & 
			{\scriptsize	 0.88} {\tiny (.02)} & 
			{\scriptsize	 0.74} {\tiny (.09)} &
			 
			  {\scriptsize	 0.76} {\tiny (.1)}& 
			  {\scriptsize	 0.86} {\tiny (.05)}&  
			  {\scriptsize	 0.73} {\tiny (.12)}\\
			  
			\small DS CQR &  
			{\scriptsize	 0.75} {\tiny (.07)} & 
			{\scriptsize	 0.88} {\tiny (.02)} & 
			{\scriptsize	 0.67} {\tiny (.09)}& 
			
			{\scriptsize	 0.79} {\tiny (.1)}& 
			{\scriptsize	 0.87} {\tiny (.04)}& 
			{\scriptsize	 0.77} {\tiny (.12)}&

			{\scriptsize	 0.73} {\tiny (.09)}& 
			{\scriptsize	 0.86} {\tiny (.03)}& 
			{\scriptsize	 0.67} {\tiny (.11)}&
			 
			{\scriptsize	 0.76} {\tiny (.11)}& 
			{\scriptsize	 0.85} {\tiny (.05)}& 
			{\scriptsize	 0.73} {\tiny (.14)}\\

            \rowcolor{green!10} \small DN NT & 
			{\scriptsize	 \textbf{0.93}} {\tiny (.02)}& 
			{\scriptsize	 \textbf{0.91}} {\tiny (.01)} & 
			{\scriptsize	 \textbf{0.95}} {\tiny (.03)}&
			  
			{\scriptsize	 \textbf{0.88}} {\tiny (.06)}& 
			{\scriptsize	 \textbf{0.90}} {\tiny (.02)}& 
			{\scriptsize	 \textbf{0.89}} {\tiny (.08)}&

			{\scriptsize	 \textbf{0.85}} {\tiny (.05)}& 
			{\scriptsize	 \textbf{0.89}} {\tiny (.02)}&  
			{\scriptsize	 \textbf{0.82}} {\tiny (.08)}&

			{\scriptsize	 \textbf{0.84}} {\tiny (.06)}&
			{\scriptsize	 \textbf{0.89}} {\tiny (.02)}&
			{\scriptsize	 \textbf{0.82}} {\tiny (.08)} 
			\\

			\midrule
			\rowcolor{black!5} \small (DN GT) & 
			{\scriptsize	 {0.93}} {\tiny (.01)}& 
			{\scriptsize	 {0.91}} {\tiny (.01)} & 
			{\scriptsize	 {0.96}} {\tiny (.02)}&
			  
			{\scriptsize	 {0.90}} {\tiny (.04)}& 
			{\scriptsize	 {0.91}} {\tiny (.01)}& 
			{\scriptsize	 {0.91}} {\tiny (.06)}&

			{\scriptsize	 {0.85}} {\tiny (.05)}& 
			{\scriptsize	 {0.89}} {\tiny (.01)}&  
			{\scriptsize	 {0.82}} {\tiny (.08)}&

			{\scriptsize	 {0.84}} {\tiny (.07)}&
			{\scriptsize	 {0.88}} {\tiny (.02)}&
			{\scriptsize	 {0.82}} {\tiny (.09)} 
			\\

            \bottomrule
		\end{tabularx}
\end{table*}

{\bf Dataset.} For this first experiment, we generate synthetic data in three steps: {\it Step 1)} Sample a DAG $\mathcal{G}$, using an Erdős–Rényi model with parameters $(d,s)$, where $d$ is the feature space dimension, and $s$ is the number of edges, which controls the sparsity of $\mathcal{G}$. {\it Step 2)} Sample Structural Equation Models (SEM), either linear or two-layered multilayer-perceptrons (MLP). {\it Step 3)} Sample the data using a topological ordering induced by $\mathcal{G}$. We consider an additive noise model, with Gaussian noise of mean $0$ and variance $1$. The parameters of the SEMs are sampled from a mixture of two uniform distributions: $\mathcal{U}(-2.5, -0.5)$ and $\mathcal{U}(0.5, 2.5)$, following \cite{Zheng2018DAGsWN}.

We then corrupt the sampled SEMs to create inconsistencies at test-time: we add Gaussian noise to the parameters of the linear SEMs and when the SEMs are MLPs, we perform the corruption on the last layer, by sampling $5$ dimensions to be corrupted. More details are included in Appendix \ref{app:exp_details}.

{\bf Methodology.} The goal of this experiment is to show that structures permit to flag inconsistencies more accurately. To show this, the methods are evaluated on their ability to flag inconsistent samples in a test set $\mathcal{D}_{\mathrm{test}}$. We consider $d=20$, $n_{\mathrm{train}} = 1000, s \in  \{10k \mid k \in [4] \}$ (controlling the sparsity in the structure), and set the same significance level $\alpha = \frac{0.1}{d}$ for every feature and every method, thereby adopting the Bonferroni correction. For every $s$, we sample $20$ DAGs, SEMs and corresponding training sets. At test time, we sample and corrupt $2$ features, by altering their corresponding SEMs. We then sample $\mathcal{D}_{\mathrm{test, corrupt}}$, with $n_{\mathrm{test}} = 10000$. In order to investigate false positives, we also sample a clean test set drawn from the same distribution as the training dataset, denoted as $\mathcal{D}_{\mathrm{test, clean}}$, with cardinality $n_{\mathrm{test}}$. The final test set is $\mathcal{D}_{\mathrm{test}} = \mathcal{D}_{\mathrm{test,corrupt}} \bigcup \mathcal{D}_{\mathrm{test,clean}}$. 

{\bf Results.} The detection task involves two classes ($1$: corrupted, $0$: non-corrupted). Hence, for each $s$, we report the $F_1$ scores, precision, and recall for the different methods. As seen in \cref{tab:exp1}, incorporating structures is consistently useful, but it is especially useful in sparse settings ($s \in \{10, 20\}$): DAGnosis does not take into account noise variables thanks to the structure. As a direct consequence, this leads to better detection results than DAGnosis Auto, which uses $d-1$ variables for each feature, and Data-SUITE CQR, which uses PCA as a representation. Beyond these metrics, we also compute an AUROC for each $s$ and each method, sweeping $\alpha$ across $\{\frac{0.1k}{d} \mid k \in [9]\}$. We report the results in \cref{app:auroc} and show that DAGnosis again outperforms Data-SUITE.

\textcolor{ForestGreen}{\textbf{Takeaway.}} Structures permit to model and take into account feature interactions. Incorporating the conditional independencies embodied in these structures makes it possible to specialize the sets of conditioning variables and ignore irrelevant variables. This leads to a better detection of inconsistencies compared to other representations of $\mathcal{D}_{\mathrm{train}}$ which ignore this information.

 \subsection{DAGnosis is Effective Even With Imperfect DAGs}\label{robustness}

{\bf Methodology.} 
In this experiment, we investigate the ability of DAGnosis to flag inconsistencies using imperfect DAGs. For that, we consider a high-dimensional synthetic setting ($d=100)$ where we either corrupt the ground-truth DAG, or learn it with a structure learner. We generate $k=5$ DAGs, with $s=50$. We consider a list of Structural Hamming Distances (SHD), namely [10, 20, 30, 40]. For each of these SHD values, we sample 5 corrupted DAGs with the given SHD from the ground-truth DAG. This mechanism directly mimics misspecifications of the DAG of various strengths, where a high SHD indicates a high misspecification.
 We also learn the DAG with the structure learner DAGMA \citep{bello2022dagma}, illustrating the flexibility of DAGnosis in the way the structure is learnt.
Finally, we sample the data with MLP SEMs, as in Section \ref{inconsistency}.

{\bf Results.} F1 score, precision and recall for the inconsistency detection task are reported in Table \ref{table:robustness}. For each SHD, we average the results over the $k$ misspecified DAGs which we sampled. These results show that DAGnosis outperforms Data-SUITE and achieves good performance even when the input DAG does not match exactly the ground-truth DAG, highlighting its robustness. We report additional results in high-dimension ($d=200$) in \cref{highdim_appendix}.

\begin{table}[t]
	\centering
	\caption{\textbf{Robustness of DAGnosis.} We report the detection metrics of the different methods, for $d=100$. DAGnosis (DN) is robust to misspecifications of the DAG, when the DAG is either corrupted or learnt with a structure learner such as DAGMA.}
 \label{table:robustness}
	\vspace{0.2em}
	\label{tab:auroc}
	\begin{tabularx}{\linewidth}{c|C|C|C}
		\toprule
			Method &  F1-score & Precision & Recall \\
        \toprule
         Data-SUITE &
        {\footnotesize	 0.49} {\tiny (.23)}& 
        {\footnotesize	 0.81} {\tiny (.07)}& 
        {\footnotesize	 0.4} {\tiny (.27)}\\
        \midrule
         DN SHD 10 & 
        {\footnotesize	 0.70} {\tiny (.1)}& 
        {\footnotesize	 0.85} {\tiny (.04)}& 
        {\footnotesize	 0.63} {\tiny (.11)}\\
        DN SHD 20 & 
        {\footnotesize	 0.73} {\tiny (.1)}& 
        {\footnotesize	 0.86} {\tiny (.04)}& 
        {\footnotesize	 0.67} {\tiny (.11)}\\
        DN SHD 30 & 
        {\footnotesize	 0.69} {\tiny (.1)}& 
        {\footnotesize	 0.85} {\tiny (.04)}& 
        {\footnotesize	 0.62} {\tiny (.11)}\\
        DN SHD 40 & 
        
        {\footnotesize	 0.63} {\tiny (.1)}& 
        {\footnotesize	 0.83} {\tiny (.04)}& 
        {\footnotesize	 0.55} {\tiny (.11)}\\
        DN DAGMA & 
        
        {\footnotesize	 \textbf{0.79}} {\tiny (.1)}& 
        {\footnotesize	 \textbf{0.89}} {\tiny (.03)}& 
        {\footnotesize	 \textbf{0.72}} {\tiny (.13)}\\
        \bottomrule
        
	\end{tabularx}
	
\end{table}

\textcolor{ForestGreen}{\textbf{Takeaway.}}   DAGnosis is robust to misspecifications of the DAG and can operate in high-dimensional setups.

\subsection{DAGnosis Unlocks Localization}

{

\paragraph{Methodology.} Having demonstrated that  DAGnosis flags inconsistencies accurately in \cref{inconsistency}, we now show that structures enable the {\it localization} of such inconsistencies, which is impossible with Data-SUITE. 
 We consider a synthetic setup, with $d = 4$. The DAG used to generate the data is the chain $X_1 \rightarrow X_2 \rightarrow X_3 \rightarrow X_4$ and the SEMs are MLPs. We only corrupt the last feature  in the topological ordering, which is $X_4$, making the inconsistencies $\textit{localized}$. Hence we desire to flag inconsistent samples solely on this feature.

\paragraph{Results.} We compute the average number of flagged features for the samples deemed inconsistent by Data-SUITE (DS) and DAGnosis (DN), with $n_{\mathrm{test}} = 10000$, denoted by $a_{\mathrm{DS}}$ and $a_{\mathrm{DN}}$. We obtain $a_{\mathrm{DS}} = \textbf{3.48}$ and $a_{\mathrm{DN}} = \textbf{1.05}$. DAGnosis is significantly more precise when flagging inconsistent samples, as it most often localizes and flags the inconsistencies only on $X_4$, while Data-SUITE flags nearly all the features on average.
}

\textcolor{ForestGreen}{\textbf{Takeaway.}} DAGnosis unlocks localization and flags inconsistencies where they happen. Hence, DAGnosis can answer two questions: is there an inconsistency? If so, \emph{where}? This property stems from its nature, since it uses {\it conditionals by design}. On the contrary, Data-SUITE can only answer one question: is there an inconsistency?

 \subsection{Reliable Downstream Performance} \label{reliable}

\begin{figure}[t]
\begin{subfigure}[b]{0.5\textwidth}
\begin{tikzpicture}
    \begin{axis}[
        xlabel= Amount of women with low income in $\mathcal{D}_{\mathrm{test}}$,
        xticklabels={0k, 2k, 4k, 6k, 8k},
        xtick={0, 2, 4, 6, 8},
        ylabel style={text width=2.5cm},
        ylabel=Accuracy on {\tiny$\mathcal{D}_{\mathrm{test}}^{(k)} \setminus \mathcal{D}_{\mathrm{flagged}}^{(k)}$},
        ylabel near ticks,
        xlabel near ticks,
        ymajorgrids=true,
        grid style=dashed,
        height=12em,
        width=\linewidth,
        legend style = {nodes={scale=0.7, transform shape}},
    ]
      \addplot[mark=*,
        blue,
        error bars/.cd, y dir=both, y explicit,] plot coordinates {
        (0,0.810971992535196) +=(0,0.0023580014665824595) -=(0,0.0023580014665824595)
        (1,0.7800346074038114) +=(0,0.0016327000824476977) -=(0,0.0016327000824476977)
        (2,0.7519601372889071) +=(0,0.0022446877402472654) -=(0,0.0022446877402472654)
        (3,0.7258779249956422) +=(0,0.0032878732240469544) -=(0,0.0032878732240469544)
        (4,0.7019517012010735) +=(0,0.004496449385048145) -=(0,0.004496449385048145)
        (5,0.6788686452905379) +=(0,0.005531807052926279) -=(0,0.005531807052926279)
        (6,0.6576804661917535) +=(0,0.006197766338192609) -=(0,0.006197766338192609)
        (7,0.6375284692334036) +=(0,0.006823392355489135) -=(0,0.006823392355489135)
        (8,0.6194665501362293) +=(0,0.007415230178285221) -=(0,0.007415230178285221)
        (9,0.6015745202153753) +=(0,0.0077252918954368545) -=(0,0.0077252918954368545)
    };
    \addplot[mark=*,
        red,
        error bars/.cd, 
        y dir=both, 
        y explicit] plot coordinates {
        (0,0.8110026092376625) +=(0,0.001693924768780603) -=(0,0.001693924768780603)
        (1,0.7756532715217895) +=(0,0.002916451477767079) -=(0,0.002916451477767079)
        (2,0.7427212101481627) +=(0,0.004066933432070744) -=(0,0.004066933432070744)
        (3,0.711782779171014) +=(0,0.004990848142717996) -=(0,0.004990848142717996)
        (4,0.6839248716451767) +=(0,0.005582797260053129) -=(0,0.005582797260053129)
        (5,0.6580259060863474) +=(0,0.006687984295369714) -=(0,0.006687984295369714)
        (6,0.6342617399462591) +=(0,0.007378761652791072) -=(0,0.007378761652791072)
        (7,0.6115753314934105) +=(0,0.007757035501633964) -=(0,0.007757035501633964)
        (8,0.5907212406039469) +=(0,0.008143727237571357) -=(0,0.008143727237571357)
        (9,0.5715327297579289) +=(0,0.00894570863562252) -=(0,0.00894570863562252)
    };
    \addplot[mark=*,
        brown,
        error bars/.cd, 
        y dir=both, 
        y explicit] plot coordinates {
        (0,0.8137971698113209) +=(0,0.0017575309106972793) -=(0,0.0017575309106972793)
(1,0.7637604525332022) +=(0,0.0016494682624744) -=(0,0.0016494682624744)
(2,0.719520389249305) +=(0,0.0015539244567239975) -=(0,0.0015539244567239975)
(3,0.6801248357424442) +=(0,0.0014688431789796003) -=(0,0.0014688431789796003)
(4,0.6448193521594684) +=(0,0.0013925950903697659) -=(0,0.0013925950903697659)
(5,0.612998420844848) +=(0,0.0013238724743823212) -=(0,0.0013238724743823212)
(6,0.5841704288939051) +=(0,0.0012616136108391607) -=(0,0.0012616136108391607)
(7,0.5579320876751707) +=(0,0.0012049475305822646) -=(0,0.0012049475305822646)
(8,0.5339494497936725) +=(0,0.0011531530184355143) -=(0,0.0011531530184355143)
(9,0.5119436201780415) +=(0,0.0011056277539104888) -=(0,0.0011056277539104888)
    };
    
    \legend{DAGnosis\\Data-SUITE\\No flagging\\}
    
    \end{axis}
\end{tikzpicture}
\caption{Downstream accuracy for non-flagged samples}

\label{fig:adult_accuracy}
\end{subfigure}
\hfill
\begin{subfigure}[b]{0.4\textwidth}

    \hspace*{-2.5mm}
    \centering
    \begin{tikzpicture}[scale=1]
        \begin{axis} 
        [
            ybar, 
            bar width=0.5cm, height=5cm, width=6.5cm,
            enlarge x limits=.9,
            ylabel= Prop. of $\mathcal{D}_{\mathrm{test}}^{(5)}$ flagged,
            symbolic x coords={Women,Men},
            ymax=0.13,
            xtick=data,
            ytick={.04, .06, .08, .10, .12},
            yticklabels={.04, .06, .08, .10, .12},
            legend style={nodes={scale=0.7, transform shape}}
        ]
        \addplot[style={red,fill=red!30, draw=red,mark=none}, error bars/.cd, y dir=both, y explicit,]
            coordinates {(Women, 0.08120805369127516) +=(0,0.008920411806355756) -= (0,0.008920411806355756)
            (Men, 0.04165021713383339) +=(0,0.002360126654677916) -= (0,0.002360126654677916)
            };
        \addplot[style={blue,fill=blue!30, draw=blue,mark=none}, error bars/.cd, y dir=both, y explicit,]
            coordinates {(Women,0.1068298460323727) += (0, 0.008403820311632153) -= (0, 0.008403820311632153)
            (Men,0.03459336754836163) += (0,0.0024748681410712596) -= (0,0.0024748681410712596)
            };
        \legend{Data-SUITE, DAGnosis PC}

        \end{axis}
    \end{tikzpicture}
    
    \caption{Analysis of flagged samples}
    \label{fig:adult_example}
\end{subfigure}
\caption{{\bf (a):} Deferring prediction on $\mathcal{D}_{\mathrm{flagged}}^{(k)}$, the set of samples flagged by DAGnosis, leads to a better downstream accuracy. {\bf (b):} We report the proportion of test samples which are flagged and are women or men, for both DAGnosis and Data-SUITE (DS). DAGnosis is more accurate than DS because it flags more inconsistent samples, while flagging a similar number of men.}
\end{figure}
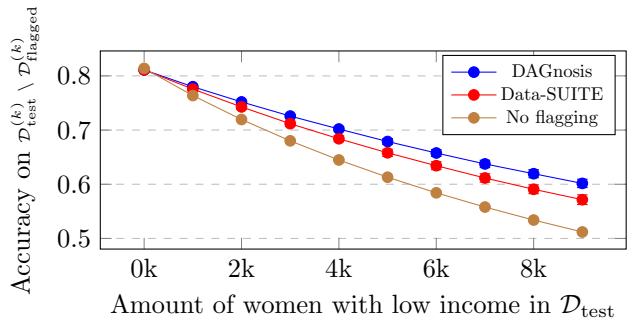
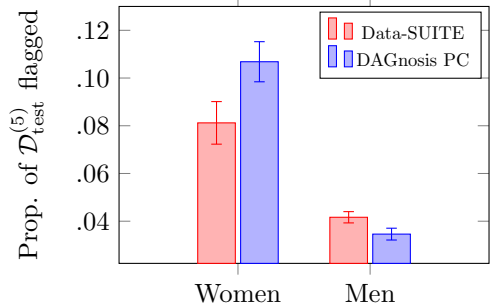

{\bf Methodology.} An important use case of DAGnosis is to ensure reliable downstream performance, where a practitioner might want to defer predictions on samples flagged as inconsistent. As an example, we use the dataset \textbf{UCI Adult income} \citep{Asuncion2007UCIML}, which captures demographic, financial and personal features, with $d = 14$.
We define specific train/test splits to control the presence of inconsistencies in the test dataset. More precisely, we split men equally in $\mathcal{D}_{\mathrm{train}}$ and $\mathcal{D}_{\mathrm{test}}$. We put women with high incomes in $\mathcal{D}_{\mathrm{train}}$.  We gradually add women with low incomes in $\mathcal{D}_{\mathrm{test}}$, with a parameter $k$ controlling the number of such samples, giving a list of $k$ test sets $\mathcal{D}^{(k)}_{\mathrm{test}}$. This motivates identifying inconsistencies in $\mathcal{D}_{\mathrm{test}}^{(k)}$ with respect to $\mathcal{D}_{\mathrm{train}}$, as we expect to flag most of the women in $\mathcal{D}_{\mathrm{test}}^{(k)}$  as inconsistent. We repeat the experiment with $5$ different seeds.

{\bf Results.} We then compute the downstream task accuracy on $\mathcal{D}_{\mathrm{test}}^{(k)} \setminus \mathcal{D}_{\mathrm{flagged}}^{(k)}$, i.e. we defer prediction on the inconsistent samples. As shown in \cref{fig:adult_accuracy}, the presence of inconsistencies has a large impact on the downstream accuracy (mean and $1.96\times$ standard error reported). DAGnosis leads to the highest downstream accuracy on $\mathcal{D}_{\mathrm{test}}^{(k)} \setminus \mathcal{D}_{\mathrm{flagged}}^{(k)}$ (when deferring prediction on inconsistent samples). 
We emphasize that this result is not a consequence of DAGnosis flagging more men in $\mathcal{D}^{(k)}_{\mathrm{test}}$ than DS. To illustrate this, we report in \cref{fig:adult_example} for $k=5$ the proportion of $\mathcal{D}_{\mathrm{test}}^{(5)}$ which is flagged and consists of women and men (mean and $1.96\times$ standard error reported).
 We conclude that both methods flag a similar amount of men. However, the key difference is that DAGnosis flags more women, who are inconsistent by design of the train/test split. We provide additional results for the \textbf{Credit} dataset \citep{yeh2009comparisons} in Appendix \ref{app:reliable}, similarly showing that DAGnosis enables reliable downstream performance.
 
\textcolor{ForestGreen}{\textbf{Takeaway.}}   DAGnosis informs the practitioner by flagging samples harmful for downstream tasks, for which predictions should be deferred, leading to better downstream performance.

\section{HOW TO USE DAGNOSIS STEP-BY-STEP} \label{walkthrough}
Having demonstrated DAGnosis' superior accuracy in identifying inconsistencies and enabling reliable downstream performance, we now provide an illustrative walkthrough on how DAGnosis can be valuable to practitioners investigating real-world data. To demonstrate its utility, we present a case study highlighting its ability to \textit{localize} the causes of inconsistencies.

{\bf Step 1. Dataset Construction.} Throughout this section, we use the real-world dataset \textbf{UCI Adult income}. As in \cref{reliable}, we assume access to $\Dtrain$ and test sets $\mathcal{D}_{\mathrm{test}}^{(k)}$, $k \in [9]$. For what follows, we take $\mathcal{D}_{\mathrm{test}} = \mathcal{D}_{\mathrm{test}}^{(5)}$. Our aim is to flag samples in $\mathcal{D}_{\mathrm{test}}$ which are inconsistent with respect to $\Dtrain$.

\textbf{Step 2. DAG Discovery.}
 Since no ground-truth DAG is available for this dataset, we use the PC algorithm and discover a DAG $\mathcal{G}$, prior to using DAGnosis. 
 
\textbf{Step 3. Flagging Inconsistencies.} Equipped with $\mathcal{G}$, we perform the machinery of DAGnosis detailed in \cref{sec:dagnosis}, by training the conformal predictors using $\Dtrain$ and $\mathcal{G}$. We then obtain the set $\mathcal{D}_{\mathrm{flagged}}$, i.e. the samples in $\mathcal{D}_{\mathrm{test}}$ which are flagged as inconsistent.

{

 Having identified a set of inconsistent samples $\mathcal{D}_{\mathrm{flagged}}$, a practitioner may desire to understand these inconsistencies with respect to $\mathcal{D}_{\mathrm{train}}$. DAGnosis empowers the practitioner in this regard because it brings \textit{localization}.

 \begin{figure}[ht]
    \centering
    \resizebox{\linewidth}{!}{
    
    \begin{tikzpicture}[
        card/.style = {rounded corners, fill=DarkGreen!10, draw=black, thick, anchor=south, align=left},
        vert/.style={circle, thick, draw=black, inner sep=0, minimum size=2mm}, 
        scale = 1
    ]
        \node[vert, fill=DarkGreen!25] (race) at (3, 0) {};
        \node[vert, fill=DarkGreen!25] (sex) at (1, 1) {};
        \node[vert, fill=DarkGreen!25] (age) at (-1, 1) {};
        \node[vert, fill=red!25] (country) at (-3, 0) {};
        
        \node[vert, fill=DarkGreen!25] (marital) at (2.5, -2) {};
        \node[vert, fill=black!25] (occupation) at (-2.5, -2) {};
        \node[vert, fill=DarkGreen!25] (relationship) at (0, -3) {};
        
        \begin{pgfonlayer}{bg}
            \node[card] at (race) {\tiny {\bf Race:} {\it white}};
            \node[card] at (sex) {\tiny {\bf Sex:} {\it female}};
            \node[card] at (age) {\tiny {\bf Age:} {\it 49}};
            \node[card, fill=red!10] at (country) {\tiny {\bf Country:} \textcolor{Maroon}{\it Peru}};

            \node[card, anchor=north] at (marital) {\tiny {\bf Marital stat.:} {\it married-civ}};
            \node[card, anchor=north] at (relationship) {\tiny {\bf Relationship:} {\it wife}};
            \node[card, anchor=north] at (occupation) {\tiny {\bf Occupation:} {\it Machine-op-insp.}};
        \end{pgfonlayer}

        \draw[-latex, thick] (race) -- (marital);
        \draw[-latex, thick] (race) -- (occupation);
        \draw[-latex, thick] (race) -- (relationship);

        \draw[-latex, thick] (sex) -- (marital);
        \draw[-latex, thick] (sex) -- (occupation);
        \draw[-latex, thick] (sex) -- (relationship);

        \draw[-latex, thick] (age) -- (marital);
        \draw[-latex, thick] (age) -- (occupation);
        \draw[-latex, thick] (age) -- (relationship);

        \draw[-latex, thick] (country) -- (marital);
        \draw[-latex, thick] (country) -- (occupation);
        \draw[-latex, thick] (country) -- (relationship);

    \end{tikzpicture}
    }
    
    \caption{We depict the Markov boundary for the feature \textit{Country}, which is flagged for the given example. An investigation of $\mathcal{D}_{\mathrm{train}}$ shows that this inconsistency can be traced back to the \textit{Occupation} feature.}
    \label{fig:example}
\end{figure}
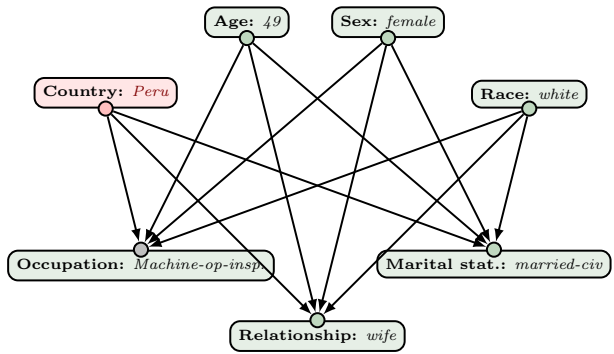
 
 \textbf{Step 4. Localizing the Inconsistencies.}  \cref{fig:example} presents an example of a sample flagged by DAGnosis (and ignored by Data-SUITE). 
 This sample is also wrongly predicted by the downstream classifier, hinting at its inconsistency.
 It is flagged by DAGnosis on the feature {\it Native-country} (Peru). We show in \cref{fig:example} the associated Markov boundary as well as the accompanying feature values. This MB is of size $6$, less than half of the total number of features ($14$). Hence DAGnosis informs the practitioner with a narrow set of variables which explain the inconsistency. 
 
 \textbf{Step 5. Gaining Understanding with $\Dtrain$.} One feature of particular interest in this MB is {\it Occupation}. We can then go back to $\Dtrain$ to understand the inconsistency, narrowing down our focus to the relationship between {\it Occupation} and {\it Country}. We notice that all the women who exhibit \textit{Occupation = Machine-op-insp.} in $\mathcal{D}_{\mathrm{train}}$ are from the \textit{United States} (which is to be expected since women in the training set have an income $>$ 50k by construction). This conflicts with the the value \textit{Peru} of the given example, which explains why  DAGnosis flags this example as inconsistent. 
 
 \textbf{Contrast with Data-SUITE.}
 On the contrary, Data-SUITE doesn't flag this sample since it loses this valuable context because of its compressive representation (e.g. PCA). Furthermore, the walkthrough shown above only makes sense in context of the DAG we learned in Step 2. Hence, Data-SUITE cannot perform Step 4 and Step 5, since it uses a compressive representation, i.e. localization is impossible. 
 
}

\section{DISCUSSION}
We have introduced DAGnosis, a data-centric method which leverages structures as representations of data in order to flag inconsistent samples at test-time. We show that structures provide specific and localized information about the consistency of each feature, leading to relevant sample-wise conclusions. We have shown experimentally that this insight provides more accurate detection of inconsistencies and ensures reliable downstream performance. DAGnosis also helps with the understanding of these inconsistencies, by localizing their causes.
Future work could build on the insight of the value of structure to advance the data-centric research agenda.

\textbf{Future Directions.} While we have focused in this work on tabular data, other data modalities might benefit from the use of structures to identify and localize inconsistencies, such as time series and natural language. This would require adapting structure discovery and conformal prediction, the two building blocks of DAGnosis.

\subsubsection*{Acknowledgements}
The authors would like to thank Tennison Liu, Hao Sun, Andrew Rashbass and the three anonymous AISTATS reviewers for useful comments on an earlier version of the manuscript. NH is funded by Illumina, JB by the W.D. Armstrong Trust, NS by the Cystic Fibrosis Trust, and JC by Aviva. This work was supported by Azure sponsorship
credits granted by Microsoft’s AI for Good Research Lab.

\bibliography{bibliography.bib}

\subsubsection*{Checklist}

 \begin{enumerate}

 \item For all models and algorithms presented, check if you include:
 \begin{enumerate}
   \item A clear description of the mathematical setting, assumptions, algorithm, and/or model. [Yes] See \cref{method}
   \item An analysis of the properties and complexity (time, space, sample size) of any algorithm. [Yes] Included in the supplementary material.
   \item (Optional) Anonymized source code, with specification of all dependencies, including external libraries. [No] Code will be released upon publication.
 \end{enumerate}

 \item For any theoretical claim, check if you include:
 \begin{enumerate}
   \item Statements of the full set of assumptions of all theoretical results. [Not Applicable]
   \item Complete proofs of all theoretical results. [Not Applicable] 
   \item Clear explanations of any assumptions. [Not Applicable]     
 \end{enumerate}

 \item For all figures and tables that present empirical results, check if you include:
 \begin{enumerate}
   \item The code, data, and instructions needed to reproduce the main experimental results (either in the supplemental material or as a URL). [Yes] Link to code is provided.
   \item All the training details (e.g., data splits, hyperparameters, how they were chosen). [Yes]
         \item A clear definition of the specific measure or statistics and error bars (e.g., with respect to the random seed after running experiments multiple times). [Yes]
         \item A description of the computing infrastructure used. (e.g., type of GPUs, internal cluster, or cloud provider). [Yes] Included in the supplementary material.
 \end{enumerate}

 \item If you are using existing assets (e.g., code, data, models) or curating/releasing new assets, check if you include:
 \begin{enumerate}
   \item Citations of the creator If your work uses existing assets. [Yes]
   \item The license information of the assets, if applicable. [Yes]
   \item New assets either in the supplemental material or as a URL, if applicable. [Yes]
   \item Information about consent from data providers/curators. [Not Applicable]
   \item Discussion of sensible content if applicable, e.g., personally identifiable information or offensive content. [Not Applicable]
 \end{enumerate}

 \item If you used crowdsourcing or conducted research with human subjects, check if you include:
 \begin{enumerate}
   \item The full text of instructions given to participants and screenshots. [Not Applicable]
   \item Descriptions of potential participant risks, with links to Institutional Review Board (IRB) approvals if applicable. [Not Applicable]
   \item The estimated hourly wage paid to participants and the total amount spent on participant compensation. [Not Applicable]
 \end{enumerate}

 \end{enumerate}

\appendix
\onecolumn
\addcontentsline{toc}{section}{Appendix}
\part{Appendix: DAGnosis: Localized Identification of Data Inconsistencies using Structures}
\mtcsetdepth{parttoc}{2} 
\parttoc

\section{ADDITIONAL DETAILS ON DAGNOSIS}
\subsection{Conformal Prediction} \label{appendix:cp}
In this section, we give additional details regarding the conformal prediction part underpinning DAGnosis.
\subsubsection{Bonferroni Correction}
In our experiments, we flag a sample $x$ if $\nu(x) > 0$, i.e. at least one of the features of $x$ is inconsistent. This is similar to a multiple hypothesis testing setup, since we are computing $\nu_{i}(x)$ for $i \in [d]$ in order to define $\nu(x)$. In order to control the FWER (Family Wise Error Rate, or the probability to make a false discovery when multiple tests are conducted),  we adopt the Bonferroni correction, which, for a desired significance level $\alpha_{0}$ consists of setting the following significance level for the individual tests: $\alpha = \frac{\alpha_{0}}{d}$. Given this, the union bound leads to $FWER \leq \sum_{i=1}^{d} \frac{\alpha_{0}}{d} = \alpha_{0}$.

\subsubsection{Calibration Step in CQR}
Suppose that lower and upper quantile estimators $\hat{q}_{i, \alpha_{lo}}$ and $\hat{q}_{i, \alpha_{hi}}$ have been trained on a proper training set $\mathcal{D}_{\mathrm{train}}^{+}$. 
The calibration step is critical in order to ensure that the confidence intervals ensure coverage guarantees, and it relies on a calibration set. 

Let us fix a calibration set $\mathcal{D}_{\mathrm{cal}} = \{ x^{(1)}, ..., x^{(n_{cal})} \}$. Following \cite{Romano2019ConformalizedQR}, and for $i \in [d]$, we compute the set of non-conformity scores 
$C_{i} = \{ E_{i,j} \mid j \in [n_{cal}] \}$ where $E_{i,j} \vcentcolon= \max \{ \hat{q}_{i, \alpha_{lo}}(x^{(j)}) -x^{(j)}_{i}, x^{(j)}_{i} -  \hat{q}_{i, \alpha_{hi}}(x^{(j)}) \}$

Next, we compute $\epsilon_{\alpha, i}$, the $(1-\alpha)(1+\frac{1}{n_{cal}})$-th empirical quantile of $C_i$, which is used to construct the prediction intervals, with $l_{i, \alpha} \vcentcolon= \hat{q}_{i, \alpha_{lo}} - \epsilon_{\alpha, i}$ and $r_{i, \alpha} \vcentcolon= \hat{q}_{i, \alpha_{hi}} + \epsilon_{\alpha, i}$.

\subsubsection{Coverage Guarantee} \label{coverageguarantee}
A compelling property of conformal prediction is the marginal coverage guarantee, which holds under the exchangeability assumption (a sequence of random variables is said to be exchangeable if any permutation of the sequence has the same joint probability distribution as the original sequence). We state the marginal coverage property stemming from this assumption:
\begin{proposition}[Marginal coverage] 
 If the data points $X^{(1)}, X^{(2)}, ..., X^{(n+1)}$ are exchangeable (with $\mathcal{D}_{cal}$  defined as the set comprising the first $n$ samples) , we have for all $i \in [d]$ that  $\mathbb{P}( X^{(n+1)}_i \in [l_{i, \alpha}(X^{(n+1)}), r_{i, \alpha}(X^{(n+1)})]) \geq 1-\alpha$ for $0 < \alpha < 1$.
 \label{validity}
\end{proposition} Note that the calibration set is implicitly used to define $l_{i,\alpha}$ and $r_{i, \alpha}$. The marginal coverage guarantee is appealing, because it permits to control a desired False Positive Rate for inconsistency detection.
\subsubsection{Quantile Regression}
In our experiments, we use a LightGBM model \citep{ke2017lightgbm} as the quantile regression backbones in CQR. In order to tune the hyperparameters of all the methods, we perform a random search with $n_{\mathrm{iter}} = 100$, to tune the number of leaves (range (10,50)), the maximum depth (range (3, 20)), the number of estimators (range (50, 300)), and the learning rate (range (0,1)). Moreover,  we perform K-fold cross-validation with $5$ folds. 
We fit the lower and upper quantile regressors using the $\alpha$-pinball loss, defined by:
\begin{align} \label{eq:pinball}
\rho_\alpha(x, x') := \begin{cases}
\alpha(x - x') & \text{if } x - x'> 0, \\
(1-\alpha)(x' - x) & \text{otherwise}
\end{cases}
\end{align}

\subsection{Inference with DAGnosis} \label{appendix:inference}

We summarize in Algorithm \ref{alg:test_cp}  how DAGnosis is used at test time to flag inconsistencies. DAGnosis outputs a confidence interval for any feature $i$ of a sample $x$. When $x_i$ does not fall inside this confidence interval, the feature $i$ of $x$ is deemed inconsistent.

\begin{algorithm} \label{alg:inference}
\caption{ {\bf -- Inference.} Using a trained DAGnosis model (cfr. \cref{alg:train_cp}), we describe how one can test the samples in $\mathcal{D}_\mathrm{test}$ for inconsistencies.}
\label{alg:test_cp}
\begin{algorithmic}
\STATE \textbf{Input:} A list of conformal predictors $\{[l_{i, \alpha}, r_{i, \alpha}] \mid i \in [d]\}$, a testing set $\mathcal{D}_{\mathrm{test}}$, an empty set of inconsistent samples $\mathcal{D}_{\mathrm{test,  incons}} = \varnothing$ 

\STATE \textbf{Output:} Updated set of inconsistent samples $\mathcal{D}_{\mathrm{test,  incons}}$ 
\FOR{$x \in \mathcal{D}_{\mathrm{test}}$ }
\FOR{$i \in [d]$}
\IF{$x_i \notin [l_{i, \alpha}(x), r_{i, \alpha}(x)]$}
\STATE Add $x$ to inconsistent samples, $\mathcal{D}_{\mathrm{test,  incons}}$ 
\ENDIF
\ENDFOR
\ENDFOR

\end{algorithmic}
\end{algorithm}

\subsection{Markov Boundaries (MB) Should be Preferred Over Parents} \label{mbbetterparents}
We now explain why DAGnosis uses $MB(X)$ rather than $Pa(X)$ to flag inconsistencies. We do so both empirically and theoretically.
\subsubsection{Empirical Demonstration}

We provide additional results comparing between using MB and Parents in a synthetic setup with MLP SEMs, for the task of detecting inconsistencies, with $d=20$, $s=20$. We report in \cref{mb_exp} the mean and standard errors of the F1 score, precision and recall for 20 runs. As we can see, using the MB leads to a more accurate detection of inconsistencies than solely using the parents of each feature.

\begin{table}[htbp]
\centering
\caption{\textbf{Markov Boundaries Capture the Relevant Information.} Comparison between conditioning on the markov boundaries (MB) or the parents nodes only, for $d=20$, $s=20$. Mean and standard error for the F1 score, precision and recall for the inconsistency detection task are reported ($\uparrow$ is better)}
\begin{tabular}{lccc}
\hline
\textbf{} & \textbf{F1} & \textbf{Precision} & \textbf{Recall} \\
\hline
\textbf{Parents} & $0.92 \pm 0.02$ & $0.91 \pm 0.01$ & $0.94 \pm 0.03$ \\
\textbf{MB} & $\textbf{0.94} \pm 0.01$ & $\textbf{0.92} \pm 0.01$ & $\textbf{0.96} \pm 0.02$ \\
\hline
\end{tabular}
\label{mb_exp}
\end{table}

\subsubsection{Theoretical Justification}

A markov boundary of $X_i$ defines the minimal set of features which contains all the information relevant to predict $X_i$. This can be stated in terms of conditional independence (Def. 4.11 in \cite{Koller2009ProbabilisticGM} and Def. 3.1 in the main paper) or equivalently mutual information, i.e. $I(X_i, S\setminus S_i \vert S_i ) = 0$, where $S_i$ is a MB of $X_i$. Furthermore, MB are minimal for this property. We emphasize that $Pa(X_i)$ does not necessarily satisfy the CI property, because $Pa(X_i)$ ignores informative nodes.
When regressing $X_i$ with the mean squared error, $I(X_i, S\setminus S_i \vert S_i ) = 0$ implies that the optimal regressor can be expressed as a function of the MB $S_i$ (see \cite{wu2011functional}).

\subsection{Differences Between DAGnosis and OOD Detection} \label{app:ood}
We now emphasize the differences between DAGnosis and OOD detectors.
\subsubsection{Most OOD Detectors are not Widely Applicable and not Tailored to Tabular Data}
Energy-based OOD detection \citep{liu2020energy} and confidence-based detection \citep{berger2021confidence} encompass a big proportion of the OOD detection litterature. However, they 
 most often rely on neural networks, yet we are interested in tabular data, where tree-based methods are prevalent. Furthermore, they also assume access to labels, which is not required by DAGnosis. Indeed, recall that DAGnosis operates at a feature level, by constructing feature-wise confidence intervals.

\subsubsection{DAGnosis Brings Localization, OOD Detectors do not}
DAGnosis permits a fine-grained analysis of samples with feature-wise confidence intervals. This fundamental novelty separates DAGnosis from OOD detectors, which traditionally adopt a generative approach based on estimating a (joint) likelihood $P(X)$. They impose a threshold on the likelihood to separate inliers from outliers. However, this does not give an explanation as to where inconsistencies happen, i.e. there is no localization. 

\subsubsection{Additional Experiment}
We compare experimentally DAGnosis to several OOD detectors tailored for the tabular domain: SUOD \citep{zhao2021suod}, Iforest \citep{liu2008isolation}, and COPOD \citep{li2020copod}.  We compute in a synthetic setup the proportion of samples flagged by each of these detectors which are also flagged by DAGnosis, for the real-world dataset \textbf{UCI Adult Income}.
We also report the downstream accuracy when we defer prediction on the samples detected as inconsistent. We report the results in Table \ref{ooddiffers} and Table \ref{ooddownstream}. 
We conclude that DAGnosis detects a more fine-grained class of inconsistent samples (i.e. in-distribution inconsistencies) thanks to localization, which are harmful for downstream tasks. As such, the work most related to ours is the SOTA Data-SUITE \citep{Seedat2022DataSUITEDI}, which also deals with ID inconsistencies. 

\begin{table}[H]
\centering
\caption{\textbf{DAGnosis Differs From OOD Detectors.} Proportion of the samples returned by the OOD detectors which are also flagged by DAGnosis (we set the thresholds such that each method flags the same number of samples)}
\begin{tabular}{cccc}
\hline
& \textbf{COPOD} & \textbf{Iforest} & \textbf{SUOD} \\
\hline
 \textbf{Overlap proportion} &0.35& 0.38 & 0.45 \\
\hline
\end{tabular}
\label{ooddiffers}
\end{table}

\begin{table}[H]
\centering
\caption{\textbf{DAGnosis Flags Harmful Inconsistent Examples.} Downstream accuracy ($\uparrow$ is better) evaluated for the samples deemed consistent by each method (complement of the inconsistent samples in the test set). Note that the thresholds of COPOD, Iforest and SUOD are set such that every method (including DAGnosis) flags the same number of inconsistencies for a fair comparison.}
\begin{tabular}{ccccc}
\hline
 & \textbf{DAGnosis} & \textbf{COPOD} & \textbf{Iforest} & \textbf{SUOD} \\
\hline
 \textbf{Downstream accuracy} &0.701 & 0.654 & 0.689 & 0.679 \\
\hline
\end{tabular}
\label{ooddownstream}
\end{table}

\newpage
\section{DETAILS ON THE EXPERIMENTAL SETUP} \label{app:exp_details}
All the experiments were run on a machine equipped with a 64-Core AMD Ryzen Threadripper and a NVIDIA RTX A4000.

\subsection{Synthetic Setup} \label{appendix:synthetic}

 \subsubsection{Generation of the Synthetic Data}
We give more details here on the setup used to generate the synthetic data in Section \ref{inconsistency}.

$\smallblacksquare$ {\bf DAG et SEM sampling} The DAGs are sampled following the Erdős–Rényi model $(d,s)$, which means that they are sampled uniformly in the set of DAGs containing $d$ nodes and $s$ edges. These DAGs define sets of parents, where $Par(i)$ is the set of parents of feature $X_i$.

{\bf Linear setting} For $i \in [d]$, we consider the SEMs defined by $X_i = W_i^{T}X_{Par(i)} + Z_i$, with $Z_i \sim \mathcal{N}(0,1)$ a gaussian noise independent of $X_{Par(i)}$ and $W_i \in \mathbb{R}^{\lvert Par(i) \rvert}$. Each dimension in the parameter $W_i$ is sampled following a mixture of the uniform distributions $\mathcal{U}(-2.5, -0.5)$ and $\mathcal{U}(0.5, 2.5)$, with weights $0.5$ and $0.5$, following \cite{Zheng2018DAGsWN}. 

{\bf MLP setting} For $i \in [d]$, we consider the SEMs defined by $X_i =  {W_i^{1}}^{T}\sigma(W_i^{2} X_{Par(i)}) + Z'_i$, with $Z'_i \sim \mathcal{N}(0,1)$ a gaussian noise independent of $X_{Par(i)}$ and $\sigma$ is the sigmoid function. Moreover, $W_i^{2} \in \mathbb{R}^{h \times \lvert Par(i) \rvert}$ and $W_i^{1} \in \mathbb{R}^{h}$, with $h=100$. Each parameter in $W_i^{1}$ and $W_i^{2}$ is sampled following a mixture of $\mathcal{U}(-2.5, -0.5)$ and $\mathcal{U}(0.5, 2.5)$, with weights $0.5$ and $0.5$.

$\smallblacksquare$ {\bf Data splitting} The training set $\mathcal{D}_{\mathrm{train}}$ is split into a proper training set $\mathcal{D}^{+}_{\mathrm{train}}$ and a calibration set $\mathcal{D}_{\mathrm{cal}}$, with $\frac{\lvert \mathcal{D}_{\mathrm{cal}} \rvert}{\lvert \mathcal{D}_{\mathrm{train}} \rvert} = 0.2$.

$\smallblacksquare$ \textbf{Generation of corruptions}
In Section \ref{inconsistency}, we corrupt the SEMs both in the linear and the MLP settings. We define these corruptions as follows.

{\bf Linear setting} We consider the perturbed parameters $W'_{i} = W_i + U_i$, with $U_i \sim \mathcal{N}(m \mathbf{1}, I)$. We take $m = 5$.  $W'_{i}$ then replaces the $W_i$ in the definition of the SEMs.

{\bf MLP setting} We consider the perturbed parameters ${W'}_i^{1} = W_i^{1} + V_i \odot M$, with $V_i \sim \mathcal{N}(m \mathbf{1}, I)$, and $M$ a random binary mask (in Section \ref{inconsistency}, we have $\sum_{i=1}^{h} M_i = 5$). Hence we perturb the last layer of the MLP. We take $m = 2$.  ${W'}_i^{1}$ then replaces $ W_i^{1}$ in the definition of the SEMs.

$\smallblacksquare$ \textbf{DAG discovery with NOTEARS}
We describe the protocol we follow to discover the DAGs. We use the differentiable structure learner NOTEARS, which comes with two variants: NOTEARS Linear \citep{Zheng2018DAGsWN} and NOTEARS MLP \citep{Zheng2020LearningSN}.

{\bf NOTEARS Linear} We use the following parameters: $\mathrm{max iter}= 100$, $h_{\mathrm{tol}} = 10^{-8}$, $\rho_{\mathrm{max}}=10^{16}$, $w_{\mathrm{threshold}}=0.3$ (adjusted when the resulting graph is not a DAG).

{\bf NOTEARS MLP} We use the following parameters: $h_{\mathrm{tol}}$ = $10^{-10}$, $\rho_{\mathrm{max}}=10^{18}$, hidden dimension of the MLP $= 10$, $\lambda_{1} = 0.01$, $w_{\mathrm{threshold}}=0.3$ (adjusted when the resulting graph is not a DAG).

\subsection{UCI Adult Income Dataset \citep{Asuncion2007UCIML}} \label{appendix:adult}
The UCI Adult income dataset was extracted from the 1994 Census bureau database. It is licensed under a Creative Commons Attribution 4.0 International. 

\subsubsection{Controlling the Inconsistencies in the Train/Test Split}
As described in Section \ref{reliable}, we construct specific train/test splits to control the amount of inconsistencies. Let us denote  $\mathcal{D}_{\mathrm{S=0,I= 0}}$ the set of samples having \textit{Sex = 0 (Women), Income $\leq$ 50k} (in what follows, \textit{Sex} is abbreviated to S, \textit{Income} to I, and the subscripts denote the values taken by these features). Furthermore, for $k \in [9]$ we consider a list of $k$ datasets, $\mathcal{D}_{\mathrm{S=0, I=0}}^{(k)} \subset \mathcal{D}_{\mathrm{S=0, I=0}}$, with $\lvert \mathcal{D}_{\mathrm{S=0,I=0}}^{(k)} \rvert = 1000k$. Given these notations, let us now define $\mathcal{D}_{\mathrm{train}} = \mathcal{D}_{\mathrm{train, S=1}} \bigsqcup  \mathcal{D}_{\mathrm{S=0,I=1}}$ and $\mathcal{D}_{\mathrm{test}}^{(k)} = \mathcal{D}_{\mathrm{test, S=1}} \bigsqcup  \mathcal{D}_{\mathrm{S=0, I=0}}^{(k)}$, with $\lvert \mathcal{D}_{\mathrm{train, S=1}} \rvert = \lvert \mathcal{D}_{\mathrm{test, S=1}} \rvert$. In a nutshell, we put women with high income in the training set, and control the number of women with low income in the test set. 

\subsubsection{DAG Discovery}
\begin{figure}
    \centering
    \includegraphics[scale = 0.3]{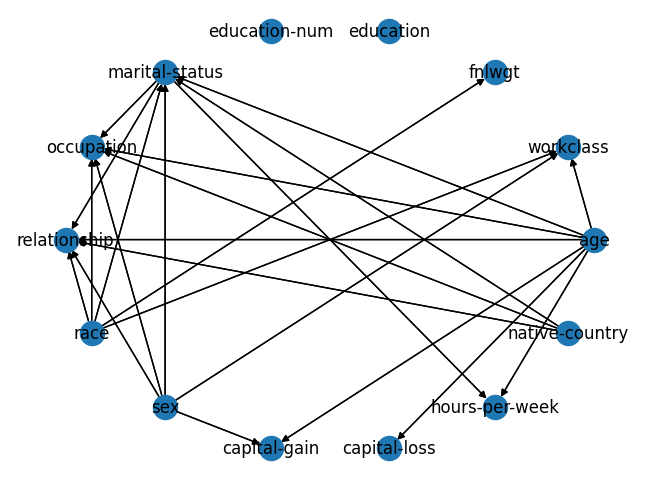}
    \caption{\textbf{Adult DAG.} DAG discovered with the PC algorithm, using $\mathcal{D}_{\mathrm{train}}$.}
    \label{fig:adult_dag}
\end{figure}

There is no ground-truth structure provided with the UCI Adult income dataset. Hence we need to discover the DAG. In order to discover the DAG, we use the PC algorithm \citep{spirtes2000causation}. We also  split the features into three different tiers,  which define a set of forbidden edges. 
\begin{itemize}
    \item Tier 1: Age, Sex, Race, Native-country
    \item Tier 2: Education, Education-num, Marital-status
    \item Tier 3: all the other features
\end{itemize}
We use the Chi-squared conditional independence test, with a significance level of 0.01, after having binned the continuous variables. The discovered DAG is depicted in Figure \ref{fig:adult_dag}. Note that we discover the DAG using only $\mathcal{D}_{\mathrm{train}}$, which is not the full dataset, to avoid any data leakage.

\subsubsection{Downstream Classifier} We consider a random forest classifier as our downstream model, with $n_{\mathrm{estimators}} = 100$. 
\newpage
\section{ADDITIONAL RESULTS}

\subsection{Sensitivity with Respect to the Discovered DAG in \ref{inconsistency}}
\textbf{Methodology.} In Section \ref{inconsistency}, we discover the DAG underlying the synthetic datasets by leveraging NOTEARS, 
and the DAG is then used by DAGnosis. A natural question one could ask is how sensitive the approach is to the discovered DAG. In order to answer this question, in the MLP SEMs setting, we plot the average Structural Hamming Distance of the discovered DAGs to the ground-truth DAG (used to generate the data), which is the minimal number of edits to go from the ground-truth DAG to the discovered DAG. This takes into account edge removals, edge additions, and edge reversals.

\begin{figure}[H]
    \centering
    \scalebox{0.7}{
\begin{tikzpicture}
 
\begin{axis}[xlabel={s},
        ylabel={Average SHD},
        ylabel near ticks,
        ymajorgrids=true,
        grid style=dashed
        ]
\addplot plot coordinates {

    (10,3.3) 
    (20,5.2) 
    (30,7.2) 
    (40,12.7)
};

\end{axis}

\end{tikzpicture}
}
\caption{\textbf{Average SHD.} We report the average Structural Hamming Distance of the discovered DAGs with respect to the ground-truth DAG, as a function of the real number of edges in this ground-truth DAG, in the MLP SEMs setting.}
\label{fig:SHD}
\end{figure}
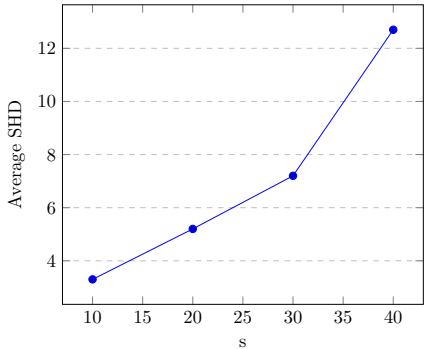
 \textbf{Results.} As we can see in \cref{fig:SHD}, the average SHD is an increasing function of the number of edges in the ground-truth DAG, which is intuitive, since a higher density implies a harder setting for DAG discovery. However, what is striking is that the results obtained by DAGnosis NOTEARS in Section \ref{inconsistency} almost match those of DAGnosis GT. This robustness result is a desideratum in  real-world settings, where the underlying DAG is often unknown and needs to be discovered.
 These results corroborate the robustness highlighted in Section \ref{robustness}, where we investigate the impact of corrupting the DAGs provided to DAGnosis.

\subsection{AUROC Results} \label{app:auroc}
\textbf{Methodology.} In addition to the F1-scores, precision and recall metrics reported in Section \ref{inconsistency}, we also compute an AUROC (Area Under the Receiver Operating Characteristic) to compare DAGnosis with Data-SUITE across different significance levels. In order to construct the ROC, we sweep the significance level $\alpha$ across $\{ \frac{0.1k}{d} \mid k \in [10] \}$. We then train the conformal estimators with each value of $\alpha$, which permits to compute at test-time a False Positive rate and a True Positive rate. The AUROC is then approximated using the trapezoidal rule. We report in \cref{tab:auroc} the mean and $1.96 \times SE$, where $SE$ denotes the standard error, for $5$ DAGs per $s \in \{10, 20, 30, 40 \}$, in the setting with MLP SEMs.

\begin{table}[htbp]
\centering
	\caption{\textbf{AUROC} We report the AUROC of the different methods, in the synthetic setting of MLP SEMs, where 5 DAGs are sampled for each $s$. DAGnosis NOTEARS consistently outperforms the two baselines, highlighting the importance of structures}
  \label{table:auroc}
	\vspace{0.2em}
\begin{tabular}{ccccc}
\hline
 & $s=10$ & $s=20$ & $s=30$ & $s=40$ \\
\hline
 \textbf{ DN Auto} &
        {\footnotesize	 0.93} {\tiny (.01)}& 
        {\footnotesize	 0.94} {\tiny (.01)}& 
        {\footnotesize	 0.94} {\tiny (.00)}& 
        {\footnotesize	 0.91} {\tiny (.01)}
        \\
        \textbf{Data-SUITE CQR} & 
        {\footnotesize	 0.93} {\tiny (.01)}& 
        {\footnotesize	 0.95} {\tiny (.00)}& 
        {\footnotesize	 0.89} {\tiny (.03)}& 
        {\footnotesize	 0.91} {\tiny (.02)}
        \\
        \rowcolor{green!10}\textbf{ DN (NT)} & 
        {\footnotesize	 0.95} {\tiny (.00)}& 
        {\footnotesize	 0.96} {\tiny (.00)}& 
        {\footnotesize	 0.95} {\tiny (.00)}& 
        {\footnotesize	 0.94} {\tiny (.01)}
        \\
        \hline
\end{tabular}
\label{highdimresults}
\end{table}
\textbf{Results.} As we can see in \cref{tab:auroc}, DAGnosis NOTEARS consistently outperforms the baselines, which corroborates the findings of Section \ref{inconsistency}, and illustrates the importance of incorporating the rich information given by structures.

\subsection{High-dimensional Results} \label{highdim_appendix}

\textbf{Methodology.} We complement Section \ref{robustness} in the main paper with additional results in high-dimension. For this, we set $d=200$, which is a 10 time increase to the dimension of the data used in Section \ref{inconsistency}. We also set $s=200$ and $n=20000$, and consider DAGnosis PC versus Data-SUITE.
We report the mean and standard errors of the F1 score, precision and recall for 10 runs in \cref{highdimresults}, for the task of detecting inconsistencies, similarly to Section \ref{inconsistency}. 

\textbf{Results.} As we can see in \cref{highdimresults}, DAGnosis outperforms the SOTA Data-SUITE by a large margin in high-dimensional settings, on all the detection metrics.
\begin{table}[htbp]
\centering
\caption{\textbf{DAGnosis also Outperforms the SOTA in High Dimension.} High-dimensional results for the task of detecting inconsistencies, $d = 200$, $s= 200$, $n = 20000$. Mean and standard error for the F1 score, precision and recall for the inconsistency detection task are reported ($\uparrow$ is better)}
\begin{tabular}{cccc}
\hline
 & \textbf{F1 score} & \textbf{Precision} & \textbf{Recall} \\
\hline
\textbf{Data-SUITE} & 0.48 $\pm$ 0.11 & 0.70 $\pm$ 0.05 & 0.36 $\pm$ 0.10\\
\textbf{DAGnosis} & \textbf{0.71} $\pm$ 0.09 & \textbf{0.86} $\pm$ 0.03 & \textbf{0.61} $\pm$ 0.11 \\
\hline
\end{tabular}
\label{highdimresults}
\end{table}

\subsection{Additional Experiment for Reliable Downstream Performance} \label{app:reliable}
\textbf{Methodology.} In this experiment, we aim to show that DAGnosis enables reliable downstream performance when deferring prediction on the set of inconsistent samples, similarly to \cref{reliable}. We use the dataset \textbf{Credit}, which is a financial
default dataset from a Taiwan bank \citep{yeh2009comparisons}, with $d = 23$.
We define specific train/test splits to control the presence of inconsistencies in the test dataset. More precisely, we split men equally in $\mathcal{D}_{\mathrm{train}}$ and $\mathcal{D}_{\mathrm{test}}$. We put women with default payment in $\mathcal{D}_{\mathrm{train}}$.  We gradually add women without default payment in $\mathcal{D}_{\mathrm{test}}$, with a parameter $k$ controlling the number of such samples, giving a list of $k$ test sets $\mathcal{D}^{(k)}_{\mathrm{test}}$. This motivates identifying inconsistencies in $\mathcal{D}_{\mathrm{test}}^{(k)}$ with respect to $\mathcal{D}_{\mathrm{train}}$, as we expect to flag most of the women in $\mathcal{D}_{\mathrm{test}}^{(k)}$  as inconsistent. 
We learn the DAG using the PC algorithm.

\begin{figure}[ht]
\begin{subfigure}{0.5\textwidth}
\centering
\begin{tikzpicture}
    \begin{axis}[
        xlabel= Amount of inconsistent women in $\mathcal{D}_{\mathrm{test}}$,
        xticklabels={0, 200, 400, 600, 800},
        xtick={0, 2, 4, 6, 8},
        ylabel style={text width=2.5cm},
        ylabel=Accuracy on {\tiny$\mathcal{D}_{\mathrm{test}}^{(k)} \setminus \mathcal{D}_{\mathrm{flagged}}^{(k)}$},
        ylabel near ticks,
        xlabel near ticks,
        ymajorgrids=true,
        grid style=dashed,
        height=12em,
        width=\linewidth,
        legend style = {nodes={scale=0.7, transform shape}},
    ]
    
      \addplot plot coordinates {
        (0, 0.804100790513834) (1, 0.796916405685377) (2, 0.7848161328588374) (3, 0.7738011695906433) (4, 0.7629151291512916) (5, 0.7528447883477469) (6, 0.7408734602463606) (7, 0.730242825607064) (8, 0.7194432361896477) (9, 0.7088065138204414) 

    };

    \addplot plot coordinates {
       (0, 0.8024090210148642) (1, 0.7812419479515589) (2, 0.7677893137503166) (3, 0.7566758173196906) (4, 0.7436840814324258) (5, 0.7328982354363065) (6, 0.7215611613517373) (7, 0.711737089201878) (8, 0.7028280018544274) (9, 0.6919214970333181) 

    };

    \addplot plot coordinates {
       (0, 0.8006393001345895) (1, 0.786896095301125) (2, 0.7740885416666666) (3, 0.7616912235746316) (4, 0.7496847414880202) (5, 0.7380509000620733) (6, 0.7267726161369193) (7, 0.715833835039133) (8, 0.7052194543297746) (9, 0.6949152542372882) 
    };
    
    \legend{DAGnosis\\Data-SUITE\\No flagging\\}
    
    \end{axis}
\end{tikzpicture}
\caption{Downstream accuracy for non-flagged samples, for the Credit dataset}
\label{fig:credit_accuracy}
\end{subfigure}
\quad
\begin{subfigure}{0.5\textwidth}
\centering

    \begin{tikzpicture}[
        card/.style = {rounded corners, fill=DarkGreen!10, draw=black, thick, anchor=south, align=left},
        vert/.style={circle, thick, draw=black, inner sep=0, minimum size=2mm}, 
        scale = 0.8,
         baseline=(current bounding box.north)
    ]
        \node[vert, fill=red!25] (marriage) at (0, 5) {};
        \node[vert, fill=DarkGreen!25] (age) at (2, 3) {};
        \node[vert, fill=DarkGreen!25] (education) at (-2, 3) {};
        \node[vert, fill=DarkGreen!25] (pay_amt6) at (0, 1) {};
        
        \begin{pgfonlayer}{bg}
            \node[card, fill=red!10] at (marriage) {\tiny {\bf Marriage:} {\it Single}};
            \node[card, anchor = west] at (age) {\tiny {\bf Age:} {\it 35}};
            \node[card, anchor = east] at (education) {\tiny {\bf Education:} {\it Graduate school}};
            \node[card, anchor = north] at (pay_amt6) {\tiny {\bf Pay AMT6:} {\it 2669}};

        \end{pgfonlayer}

        \draw[-latex, thick] (marriage) -- (education);
        \draw[-latex, thick] (age) -- (marriage);
        \draw[-latex, thick] (age) -- (education);
        \draw[-latex, thick] (pay_amt6) -- (education);

    \end{tikzpicture}

    \caption{We depict the Markov boundary for the feature \textit{Marriage}, which is flagged by DAGnosis for the given example.}
    \label{fig:credit_example}
    \vspace{-10pt}
\end{subfigure}
\caption{{\bf (a):} Deferring prediction on $\mathcal{D}_{\mathrm{flagged}}^{(k)}$, the set of samples flagged by DAGnosis, leads to a better downstream accuracy. {\bf (b):} DAGnosis provides localization. This localization contrasts Data-SUITE which uses compressive representations.}
\end{figure}
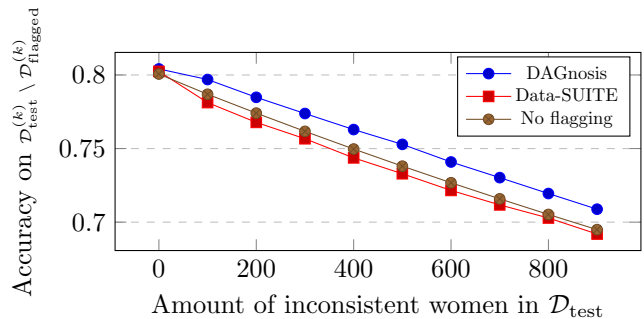
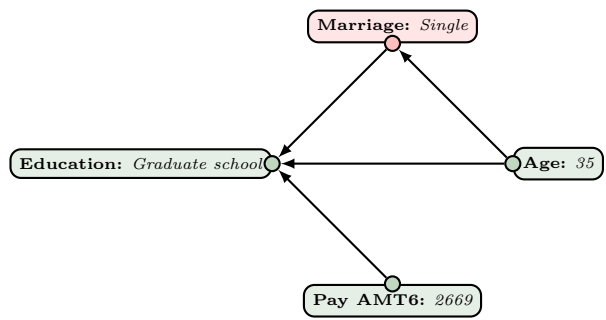

{\bf Results.} We then compute the downstream task accuracy on $\mathcal{D}_{\mathrm{test}}^{(k)} \setminus \mathcal{D}_{\mathrm{flagged}}^{(k)}$, i.e. we defer prediction on the inconsistent samples. As shown in \cref{fig:credit_accuracy}, DAGnosis flags samples which are more harmful for the downstream task than samples flagged by Data-SUITE, evidenced by the higher downstream accuracy on $\mathcal{D}_{\mathrm{test}}^{(k)} \setminus \mathcal{D}_{\mathrm{flagged}}^{(k)}$. We also show in \cref{fig:credit_example} an example of a sample flagged by DAGnosis. It illustrates the localization property inherent to our method, which contrasts Data-SUITE -- a method incapable of providing such localization because it uses compressive representations.

\newpage
\section{STRUCTURE LEARNING OVERHEAD}
In this section, we address the questions revolving around structure learning and its time/computation overhead, which is inherently defined by the structure learning method used. While structure learning is difficult in high-dimensional setups and is not an easy task in the real world, there are several reasons which make it doable as part of DAGnosis. 

{\bf Structure Learning is Conducted Once.} 
We emphasize that structure learning is a step which is taken independently from feature-wise conformal prediction. This step is conducted only once for a given training dataset, and can be done in parallel to training the model in the first place. Moreover, DAGnosis is very flexible in the way the structure is provided. Indeed, when the structure needs to be discovered, DAGnosis is completely agnostic to the structure learner.  We illustrate this with three structure learners which are the PC algorithm, NOTEARS, and DAGMA.

{\bf Recent Advances in Structure Learning.}
The flexibility in the way the structure is learnt is a key advantage of DAGnosis, because it permits to take full advantage of the recent advances in structure learning, which have made the cost of learning structures completely bearable. As an example, we refer to DAGMA \citep{bello2022dagma}, and especially Figures 4 and 5 in the corresponding paper, which illustrate the number of dimensions for which the method can be scaled up to, and the runtime. As we can see, in its linear version, this structure learner can tackle dimensions up to 1000, in a reasonable time, while maintaining good discovery accuracy.

\section{BROADER IMPACT}
The research presented in this work on the identification and handling of inconsistencies in data at deployment time holds significant broader impacts for the machine learning community. By addressing the limitations of existing data-centric methods DAGnosis brings valuable insights and advancements to the field.

One of the key broader impacts of DAGnosis is its potential to enhance the reliability and trustworthiness of machine learning models. Inconsistencies in data can significantly impact the performance of models, leading to potential errors and misinterpretations in real-world applications.  By leveraging structural interactions, DAGnosis enables more accurate conclusions in detecting inconsistencies, thereby improving the overall reliability of machine learning models in practical settings.  It is vital that inconsistency detection does not create additional bias in the real world. First, we stress that DAGnosis assesses $\Dtest$ with respect to the reference dataset $\Dtrain$, which is typically assumed to be a representative dataset of the distribution of interest. Second, DAGnosis provides a safeguard against erroneous inconsistency detection with the marginal coverage guarantee under exchangeability.

Another significant broader impact of this research is the localization of causes of inconsistencies on a DAG. Previous approaches have often lacked the ability to pinpoint the specific reasons why a sample might be flagged as inconsistent. However, DAGnosis addresses this limitation by providing detailed insights into the factors contributing to inconsistencies. This localization capability not only helps in understanding and interpreting the flagged samples but can also guide future data collection. Researchers and practitioners can use this information to refine data collection strategies, thereby improving the quality of training data.

By addressing the limitations of existing approaches, DAGnosis paves the way for more accurate and insightful data-centric conclusions.

\end{document}